\documentclass[submission, PhysCommReports]{SciPost}

\usepackage[utf8]{inputenc} 
\usepackage[T1]{fontenc} 	
\usepackage[english]{babel} 

\usepackage[bitstream-charter]{mathdesign}
\usepackage{geometry} 		
\usepackage{mathtools} 		
\usepackage{float} 			
\usepackage{graphicx} 		
\usepackage{tabularx} 		
\usepackage{booktabs} 		
\usepackage{color, xcolor} 	
\usepackage{pdfpages} 		
\usepackage{extarrows} 		
\usepackage{multirow} 		
\usepackage{multicol} 		
\usepackage{caption}
\usepackage{wrapfig}
\usepackage{enumitem} 		
\usepackage{xspace} 		
\usepackage{stackrel} 		
\usepackage{tikz} 			
\usepackage{braket} 		
\usepackage{bm} 			
\usepackage{tensor} 		
\usepackage{slashed} 		
\usepackage{siunitx} 		
\usepackage{lastpage} 		
\usepackage{cite} 			
\usepackage[normalem]{ulem} 
\usepackage{fontawesome} 	
\usepackage{tocloft} 		
\usepackage{titlesec} 		
\usepackage{doi} 			
\usepackage{hyperref} 		
\usepackage[most]{tcolorbox} 					
\usepackage[nameinlink, capitalize]{cleveref} 	
\usepackage[nottoc, notlot, notlof]{tocbibind} 	
\usepackage[ruled, vlined]{algorithm2e} 		
\usepackage{makecell}
\usepackage[makeroom]{cancel}
\usepackage{feynmf}

\makeatletter
\def\BState{\State\hskip-\ALG@thistlm}
\makeatother

\makeatletter
\@ifundefined{pdfoutput}{}{\DeclareGraphicsRule{*}{mps}{*}{}}
\makeatother

\makeatletter
\DeclareRobustCommand*{\bfseries}{%
   \not@math@alphabet\bfseries\mathbf
   \fontseries\bfdefault\selectfont
   \boldmath
}
\makeatother

\DeclareSymbolFont{usualmathcal}{OMS}{cmsy}{m}{n}
\DeclareSymbolFontAlphabet{\mathcal}{usualmathcal}

\SetArgSty{textnormal}
\SetKwComment{Comment}{{\small\#}~}{}
\SetCommentSty{mycommfont}

\setitemize{itemsep=0pt, parsep=0pt} 				
\setenumerate{itemsep=0pt, parsep=0pt} 				
\setitemize{itemsep=2pt,topsep=2pt,parsep=0pt,partopsep=0pt,leftmargin=*}
\setenumerate{itemsep=0pt,topsep=2pt,parsep=0pt,partopsep=0pt,labelindent=3pt,leftmargin=*}
\newlist{todolist}{itemize}{2}
\setlist[todolist]{label=$\square$}
\usepackage{pifont}

\usepackage{amsmath}
 
\usepackage{amsthm} 		
\theoremstyle{definition}

\definecolor{red_cb}{HTML}{e41a1c}
\definecolor{blue_cb}{HTML}{377eb8}
\definecolor{green_cb}{HTML}{4daf4a}
\definecolor{purple_cb}{HTML}{984ea3}
\definecolor{orange_cb}{HTML}{ff7f00}

\crefname{section}{Sec.}{Secs.}
\Crefname{section}{Section}{Sections}

\definecolor{EmeraldGreen}{HTML}{1ea78d}
\definecolor{EnglishRed}{HTML}{b02427}
\hypersetup{
	pdftitle={UQVERaIPHY},
	pdfauthor={Haussmann et al.},
	colorlinks=true, 			
	linkcolor={red!50!black}, 	
	citecolor=EmeraldGreen, 	
	urlcolor=EmeraldGreen 	
}

\newcommand{\eg}{\text{e.g.}\;}
\newcommand{\ie}{\text{i.e.}\;}

\newcommand{\Betadist}{\mathcal{B}\textit{eta}}

\newcommand{\kl}{\text{KL}}

\newcommand{\mwith}{\text{with}}
\newcommand{\mand}{\text{and}}

\newcommand{\qqquad}{\qquad\quad}

\def\d{\mathrm{d}}

\newcommand\one{\leavevmode\hbox{\small1\normalsize\kern-.33em1}}

\newcommand{\loss}{L} 	
\newcommand{\divf}{d}

\newcommand{\arXiv}[2][]{%
	\ifthenelse{\equal{#1}{}}%
	{\href{http://arxiv.org/abs/#2}{arXiv:#2}}%
	{\href{http://arxiv.org/abs/#2}{arXiv:#2~[#1]}}}

\def\slashchar#1{\setbox0=\hbox{$#1$}           
   \dimen0=\wd0                                 
   \setbox1=\hbox{/} \dimen1=\wd1               
   \ifdim\dimen0>\dimen1                        
      \rlap{\hbox to \dimen0{\hfil/\hfil}}      
      #1                                        
   \else                                        
      \rlap{\hbox to \dimen1{\hfil$#1$\hfil}}   
      /                                         
   \fi}

\newcommand{\tikznode}[2]{%
\ifmmode%
\tikz[remember picture,baseline=(#1.base),inner sep=0pt] \node (#1) {$#2$};%
\else
\tikz[remember picture,baseline=(#1.base),inner sep=0pt] \node (#1) {#2};%
\fi}

\def\mathswitchr#1{\relax\ifmmode{\mathrm{#1}}\else$\mathrm{#1}$\xspace\fi}
\def\mathswitch#1{\relax\ifmmode#1\else$#1$\xspace\fi}

\begin{document}

\begin{center}
{\Large \textbf{\color{scipostdeepblue}{Uncertainty in Physics and AI:\\[0.3ex]
Taxonomy, Quantification, and Validation}}} \\[0.3cm]
{\large Part of the VERaiPHY Initiative}
\end{center}

\begin{center}
Manuel Hau{\ss}mann\textsuperscript{$\star$,1},
Ramon Winterhalder\textsuperscript{$\star$,2}, and
Maria Ubiali\textsuperscript{$\circ$,3}
\end{center}

\begin{center}
{\bf 1} Department of Mathematics and Computer Science, University of Southern Denmark \\
{\bf 2} TIFLab, Universit\`a degli Studi di Milano \& INFN Sezione di Milano, Italy\\
{\bf 3} DAMTP, University of Cambridge, Wilberforce Road, Cambridge CB3 0WA, UK \\
\end{center}

\begin{center}
$\star$ {\small Leading authors}\qquad
$\circ$ {\small Advisor}
\end{center}


\section*{\color{scipostdeepblue}{Abstract}} 
{\bf 
Reliable uncertainty quantification is essential for the use of machine learning in physics, where scientific discoveries depend on validated probabilistic statements. %
We provide a structured overview of uncertainty quantification in ML for physics, introducing a unified taxonomy of uncertainty and clarifying the interpretation of predictive and inference uncertainties across frequentist and Bayesian frameworks. 
We discuss principled validation tools, including coverage, calibration, bias tests, and proper scoring rules, and illustrate them with simple regression and classification examples. 
}

\vspace{1pt}
\noindent\rule{\textwidth}{1pt}
\tableofcontents\thispagestyle{fancy}
\noindent\rule{\textwidth}{1pt}
\vspace{1pt}

\clearpage
\section{Introduction}
\label{sec:intro}

Uncertainty quantification is an important cornerstone of both physics~\cite{stat_phys_2006,Trotta:2008qt,hogg2010astro,Cowan:2010js,Trotta:2017wnx} and machine learning (ML)~\cite{Kendall2017_uq,MLgenUQ1,MLgenUQ2,MLgenUQ3,Abdar_2021,PHYgenUQ1}.
In physics, reliable uncertainty estimates are indispensable: without them, a result is scientifically incomplete. 
They determine the credibility of measurements, underpin the combination of experimental and theoretical results, dictate the strength of discovery claims, and govern how tightly 
any physical parameter can be constrained. 
In particle physics, statistical significance thresholds benchmark the reporting of new particles or rare processes, and systematic uncertainties from detector effects or theoretical modelling often dominate the final error budget. 
In astrophysics and cosmology, uncertainty estimates play an equally central role in inferring cosmological parameters, characterising sources, and performing model comparisons.

In machine learning, uncertainty estimates have become equally important. 
Modern AI systems are increasingly deployed in safety-critical applications ranging from medical diagnostics to autonomous driving,
where trustworthy predictions cannot be reduced to a single point estimate~\cite{Kompa2021_uncertainty_medical,Loftus2022_uncertainty_healthcare,Araujo2024_autonomous_driving_uncertainty}. 
Instead, decision-making depends on understanding how confident a model is in its outputs. 
Calibration of predictive probabilities, quantification of model uncertainty, and guarantees on generalization performance are therefore central to ensuring that ML methods can be applied responsibly~\cite{pmlr-v70-guo17a}. 

When ML methods are used in physics, trustworthy uncertainty estimates are equally important: scientific conclusions must be accompanied by statistically validated uncertainty statements, and if ML-based estimates are unreliable, the conclusions drawn from them should be called into question.
There already exist applications in physics where ML methods provide not only predictions but also associated uncertainty estimates, 
including examples in particle physics~\cite{Forte:2002fg,Ball:2008by,Englert:2018cfo,Bollweg:2019skg,Carrazza:2019mzf,Nachman:2019dol,Kasieczka:2020vlh,Araz:2021wqm,Bellagente:2021yyh,Ghosh:2021hrh,Ghosh:2021roe,NNPDF:2021njg,NNPDF:2021uiq,Badger:2022hwf,Chen:2022pzc,Butter:2022vkj,Fanelli:2023lmp,Heimel:2023mvw,ATLAS:2024rpl,Kriesten:2024ist,Benato:2024lnj,Huetsch:2024quz,Bieringer:2024nbc,Bahl:2024gyt,Benevedes:2025nzr,Khot:2025kqg,Bahl:2025xvx,Bendavid:2025urn,Araz:2025vuw,Costantini:2025wxp,Costantini:2025agd,Beccatini:2025tpk,Benato:2025rgo,Barontini:2025lnl,Bahl:2026qaf}, 
cosmology~\cite{Padilla:2019mgi,Rover:2024pvr,Herold:2024enb,Drabicki:2026rgn,Soriano:2026mwu,Leterme:2026cqj,Dai:2026jme} and astrophysics~\cite{Butter:2021mwl,Malz:2024zjd, Ashton:2024wae,Sun:2024ywb,Perez-Romero:2025gwg,Liu:2026nhu,walls2026montecarloconformalprediction,Mendes:2026fsj,
hong2026uncertaintyawaresolarflareregression}.
While these developments demonstrate the practical relevance of ML-based uncertainty quantification in physics, a unified statistical interpretation of such uncertainty 
estimates -- and clear guidelines for their validation -- is often still lacking across disciplines.

To clarify these issues, it is useful to begin by examining what we mean by the term \emph{uncertainty}. 
One may consider uncertainty about inferred quantities such as model parameters or functions (\emph{inference uncertainty}), or uncertainty about future observations (\emph{predictive uncertainty}). 
While these notions are related -- for instance through the posterior predictive distribution in Bayesian inference -- they obey different validation criteria and correspond to different statistical questions. 
In particular, coverage, calibration, and bias diagnostics must always be interpreted with respect to the object under consideration.
In this contribution we primarily focus on \emph{predictive} uncertainties. This choice reflects that predictive uncertainty is the quantity most directly accessible to model-agnostic validation: given a new observation, one can empirically check whether it falls within the predicted interval, without reference to unobservable true parameters. 
Inference uncertainty -- \eg the uncertainty on quantities such as $\alpha_\mathrm{S}$, $m_\mathrm{t}$, or parton distribution functions in particle physics, or cosmological parameters such as $H_0$ and  $\Omega$ -- is equally central to physics but is the primary focus of a companion VERaiPHY contribution on parameter estimation~\cite{VERaiPHY_inference}; here it enters insofar as it propagates into predictive uncertainty via marginalisation.

Despite the central role of uncertainty in both physics and statistics, the terminology used to describe it remains inconsistent across communities. 
Physicists traditionally distinguish between \emph{statistical} and \emph{systematic} uncertainties, while the statistics and ML literature more often refers to \emph{aleatoric} and \emph{epistemic} uncertainties. 
These vocabularies overlap but are not synonymous, and careless translation between them may cause confusion. 
For instance, statistical fluctuations from limited data can leave model parameters under-constrained (an epistemic effect), which may be further amplified by aleatoric randomness in the data. 
Likewise, systematic uncertainties may stem either from irreducible noise sources (aleatoric) or from potentially reducible modelling assumptions (epistemic). One of our goals in this article is therefore to provide a clear taxonomy that reconciles these perspectives and clarifies their relationships.

A further challenge lies in the validation of uncertainty estimates. 
In classical statistics, well-defined procedures exist to validate uncertainties, such as confidence intervals and hypothesis tests in the frequentist framework, or posterior distributions and credible intervals in the Bayesian framework. 
Modern machine learning methods for uncertainty quantification vary in their theoretical grounding. Some approaches, such as conformal prediction, provide finite-sample coverage guarantees under minimal assumptions~\cite{vovk2005algorithmic,shafer2007tutorialconformalprediction}. 
Others, such as deep ensembles or Bayesian neural networks with approximate inference, rely more heavily on empirical validation 
and heuristic calibration~\cite{lakshminarayanan2017}. 
Assessing whether the uncertainties produced by such models are meaningful requires careful use of statistical diagnostics such as coverage tests, pull distributions, calibration curves, and proper scoring rules~\cite{Gneiting01032007}.
These validation concepts are inherently probabilistic statements whose interpretation depends on what is treated as random -- for instance the training data, model parameters, or future observations.
Different choices lead to distinct notions of coverage and calibration, such as marginal versus conditional guarantees, or inference versus predictive uncertainty.
Understanding how to apply and interpret these diagnostics in physics contexts is therefore a key objective of this work.
Our focus is not the accuracy of point predictions, but the validity 
of the \emph{uncertainty statements} that accompany them: scientific 
applications require statistically meaningful uncertainty statements 
-- for instance prediction intervals, credible regions, or calibrated 
probability distributions -- whose consistency must be validated.

This article contributes to VERaiPHY (Validation \& Evaluation for Robust AI in PHYsics), a PHYSTAT review series establishing verification and validation standards for machine learning across particle physics, astrophysics, and cosmology~\cite{VERaiPHY_intro}.
Within this broader initiative, the present article aims to provide a structured overview of uncertainty quantification in physics-inspired machine learning workflows. 
Our scope is deliberately complementary to related VERaiPHY contributions focusing on statistical inference (parameter estimation)~\cite{VERaiPHY_inference} and hypothesis testing~\cite{VERaiPHY_hypo}. 
While statistical inference emphasizes the extraction of physical parameters from data, the focus here is on identifying sources of uncertainty, clarifying their statistical interpretation, and discussing practical tools for their quantification and validation.
Our goal is not to advocate a particular statistical paradigm or to prescribe a single \emph{best} method for uncertainty quantification. 
Instead, we aim to provide conceptual clarity, bridge terminological differences between physics and ML communities, and highlight the statistical tools that allow researchers to assess whether their uncertainty estimates can be trusted. In doing so, we hope to establish a 
common ground for physicists, statisticians, and ML researchers to build upon in future work.

The structure of the article is as follows. 
In \cref{sec:taxonomy}, we introduce a taxonomy of uncertainty and provide a glossary clarifying commonly used terminology. 
\Cref{sec:freq_vs_bayesian} contrasts frequentist and Bayesian perspectives on uncertainty and clarifies the probabilistic interpretation of uncertainty statements in each framework.
These foundations are essential for the discussion of validation diagnostics in \cref{sec:evaluation}, where we analyse tools such as coverage tests, calibration curves, and scoring rules.
In \cref{sec:examples_methods}, we illustrate these concepts through simple regression and classification examples, demonstrating commonly used approaches such as repulsive ensembles, Bayesian neural networks, Gaussian processes, and conformal prediction; evidential deep learning 
is additionally discussed for completeness. 
We conclude in \cref{sec:outlook} with a summary and outlook.

\clearpage
\section{Taxonomy of uncertainty}
\label{sec:taxonomy}

Uncertainty arises in physics and machine learning from many different sources, and its interpretation
depends strongly on the statistical framework employed.
A recurring source of confusion is that the term ``uncertainty'' is used in multiple, partly overlapping meanings. We therefore aim to keep terminology explicit by tracking
\begin{enumerate}
    \item the \emph{object} of uncertainty (inference vs.~prediction);
    \item its \emph{source} (statistical vs.~systematic);
    \item and its \emph{nature} (aleatoric vs.~epistemic).
\end{enumerate}

\subsection{Dimensions of uncertainty}

The terminology surrounding uncertainty is often inconsistent across physics, statistics, and machine learning. To provide clarity, we distinguish between two \emph{complementary} axes of uncertainty:
\begin{enumerate}
    \item \textbf{Statistical vs.~Systematic} (sources)
    \begin{itemize}
        \item \emph{Statistical uncertainty} arises from finite datasets or stochastic fluctuations in repeated experiments.
        It decreases as more data is acquired.
        \item \emph{Systematic uncertainty} arises from imperfect knowledge of measurement conditions, detector effects, or theoretical/modelling assumptions. It persists even with infinite data unless the underlying assumptions are corrected.
    \end{itemize}

    \item \textbf{Aleatoric vs.~Epistemic} (nature / reducibility)
    \begin{itemize}
        \item \emph{Aleatoric uncertainty}, or \emph{data uncertainty}, refers to randomness in the data-generating process that is treated as irreducible \emph{within the chosen modelling framework}: it cannot be decreased by collecting more data or by better optimisation, given the model class. 
        \item \emph{Epistemic uncertainty}, or \emph{model uncertainty}, refers to uncertainty due to lack of knowledge about the model,
        its parameters, or its domain of validity; in principle, it can be reduced with more data, improved modelling, or better theory input.
    \end{itemize}
    A subtle point deserves emphasis here. For many practical sources of uncertainty, the aleatoric--epistemic split depends on the chosen model class: as noted already in the engineering reliability literature~\cite{DerKiureghian2009}, what is treated as irreducible noise in one model can, in principle, be resolved into structured (epistemic) variability by a richer model that tracks additional latent degrees of freedom. Classical pseudo-randomness, such as a die roll, illustrates this: it is deterministic given enough microscopic information, and is only ``aleatoric'' relative to coarse-grained models that do not track it. In fundamental physics, however, this model-relative view has clear limits. 
    In standard quantum mechanics, randomness -- \eg in radioactive decay or measurement outcomes -- is irreducible as a result of physical law rather than modelling convention, and no widening of the hypothesis class recovers determinism. The aleatoric category therefore mixes two qualitatively different things: practical irreducibility relative to a chosen model, and fundamental irreducibility built into the underlying theory. Both behave the same way for the purposes of the decompositions that follow, but it is worth keeping the distinction in mind. See Refs.~\cite{Hullermeier2021,Gruber2023,smith2025rethinking} for extended discussions of the model-relative case.
\end{enumerate}
These two axes are not synonymous but complementary. Crucially, statistical/systematic classify \emph{sources} of uncertainty, whereas aleatoric/epistemic classify their \emph{nature}. One should therefore avoid implicitly associating statistical$\leftrightarrow$aleatoric or systematic$\leftrightarrow$epistemic, since both axes cut across each other. The interplay is summarized in Table~\ref{tab:uncertainty_dimensions}. 

\begin{table}[t!]
\centering
\begin{small}
\begin{tabularx}{0.96\linewidth}{
  l
  @{\hspace{0.8em}}
  >{\raggedright\arraybackslash}X
  @{\hspace{0.8em}}
  >{\raggedright\arraybackslash}X
}
\toprule
 & \textbf{Aleatoric (data uncertainty)}
 & \textbf{Epistemic (model uncertainty)} \\
\midrule
\textbf{Statistical}
& \emph{Not applicable} (see caption)
& Inference variability from limited data, \eg weakly constrained parameters or training instability \\
\midrule
\textbf{Systematic}
& Irreducible stochasticity in measurement processes, \eg quantum-mechanical randomness or intrinsic detector noise
& Imperfect modelling assumptions or missing theory input, \eg missing higher-order corrections \\
\bottomrule
\end{tabularx}
\end{small}
\caption{Two-dimensional classification of the \emph{source} and \emph{nature} of uncertainty. The third dimension -- the \emph{object} of uncertainty (inference vs.\ predictive) -- cuts across both axes and is discussed separately below. 
The (statistical, aleatoric) cell is intentionally left empty: aleatoric uncertainty refers to randomness in the data-generating process itself, which by construction does not decrease with sample size, so a ``statistical aleatoric'' uncertainty has no coherent meaning. What is sometimes informally called statistical aleatoric uncertainty -- \eg the shrinking error bar on an inferred Poisson rate as $N$ grows -- is in fact statistical \emph{epistemic} uncertainty, since it concerns our estimate of a parameter rather than the per-event randomness itself.}
\label{tab:uncertainty_dimensions}
\end{table}

\subsubsection*{Object of uncertainty: inference vs.\ prediction}

The axes above classify uncertainty by \emph{source} (statistical vs systematic) and \emph{nature} (aleatoric vs epistemic). A separate and often conflated distinction concerns the \emph{object} of uncertainty:
\emph{inference uncertainty} refers to uncertainty about inferred model quantities (parameters or functions),
whereas \emph{predictive uncertainty} refers to uncertainty about future or unobserved observables.
Let $\theta$ denote model parameters (or more generally a hypothesis) and let $\mathcal D$ denote the observed dataset. In Bayesian inference, \emph{inference uncertainty} is represented by the posterior
\begin{equation}
  p(\theta |\mathcal D),
\end{equation}
while in frequentist inference it is represented by the sampling variability of an estimator $\hat\theta(\mathcal D)$ and the associated confidence regions. In both cases, inference uncertainty lives in \emph{parameter or hypothesis space}. 
\emph{Predictive uncertainty} concerns a future outcome $y_\ast$ at input $x_\ast$ and is represented by a predictive distribution. In Bayesian form, the posterior predictive distribution is given by
\begin{equation}
p(y_\ast|x_\ast,\mathcal D)
= \int \d\theta\; p(y_\ast|x_\ast,\theta)\,p(\theta|\mathcal D).
\label{eq:posterior_predictive}
\end{equation}
This identity makes the link to aleatoric vs.\ epistemic explicit: the conditional distribution $p(y_\ast|x_\ast,\theta)$ encodes irreducible randomness in outcomes (aleatoric uncertainty), while uncertainty
over $\theta$ encoded in $p(\theta|\mathcal D)$ (epistemic uncertainty) propagates additional variability in predictions. A convenient framework-neutral expression is the law of total variance,
\begin{equation}
\mathrm{Var}(y_\ast|x_\ast,\mathcal D)
=
\underbrace{\mathbb E_{\theta|\mathcal D}\!\left[\mathrm{Var}(y_\ast|x_\ast,\theta)\right]}_{\text{aleatoric}}
+
\underbrace{\mathrm{Var}_{\theta|\mathcal D}\!\left(\mathbb E[y_\ast|x_\ast,\theta]\right)}_{\text{epistemic}}.
\label{eq:total_variance_decomp}
\end{equation}
This shows that epistemic uncertainty, encoded at the inference level 
in $p(\theta|\mathcal{D})$, contributes additional spread to predictions 
through marginalization over $\theta$, whereas aleatoric uncertainty 
enters directly as irreducible spread in outcomes conditional on the model.
In particular, aleatoric uncertainty manifests primarily in \emph{predictive} uncertainty: even with perfect knowledge of $\theta$, the spread in $p(y_\ast|x_\ast,\theta)$ remains. 
Inference uncertainty, by contrast, is primarily epistemic in nature -- even though its magnitude is modulated by the aleatoric noise in the observations -- and vanishes in the large-data limit under a well-specified model.

\subsection{Decomposition of epistemic uncertainty}
\label{sec:epistemic_learning}

To make the distinction between inference and prediction more explicit in a learning setting, it is useful to disentangle the
different distributions that typically appear in supervised learning. 

Let $p(y| x)$ denote the (unknown) \emph{data-generating} conditional distribution.
A learning algorithm explores a restricted hypothesis space $\mathcal{H}$ of representable conditionals (\eg a neural network family).
Training on a dataset $\mathcal D_N$ of size $N$ produces an estimator $\hat p_{N,\alpha}(y| x)$, where $\alpha$ denotes sources of algorithmic variability such as random initialization, stochastic optimisation noise, mini-batch order, and early stopping.
To define suitable reference points in $\mathcal H$, we introduce
\begin{enumerate}[label=(\roman*)]
    \item an \emph{idealised finite-sample} solution $\hat p_N(y|x)$ as an empirical-risk minimiser for $\mathcal D_N$;
    \item and a \emph{best-in-class} conditional $\hat p(y|x)$ as the corresponding population-risk minimiser within $\mathcal H$.
\end{enumerate}
These definitions identify three distinguished elements in hypothesis space: the algorithm-dependent outcome $\hat p_{N,\alpha}$, the
idealised finite-sample target $\hat p_N$, and the best-in-class limit $\hat p$.
If $\mathcal H$ is parametric, $\mathcal H=\{p(y|x,\theta):\theta\in\Theta\}$, these correspond to $\hat p_{N,\alpha}\equiv p(y|x,\hat\theta_{N,\alpha})$, ${\hat p_N \equiv p(y|x,\hat\theta_N)}$, and $\hat p\equiv p(y|x,\theta_\text{opt})$, respectively.
These objects suggest a \emph{conceptual} decomposition of inference-level epistemic uncertainty illustrated in \cref{fig:hypothesis_space}. The split is meant to highlight dominant sources and is not strictly separable in practice: changing dataset size or composition may help or hinder optimisation stability, and model misspecification can modify both finite-$N$ and
asymptotic behaviour. We quantify gaps using a generic discrepancy measure $\divf(\cdot,\cdot)$ between conditional distributions (\eg KL divergence, Wasserstein distance, total variation, etc.) and decompose into:

\begin{figure}[t!]
\centering
\includegraphics[width=0.5\linewidth]{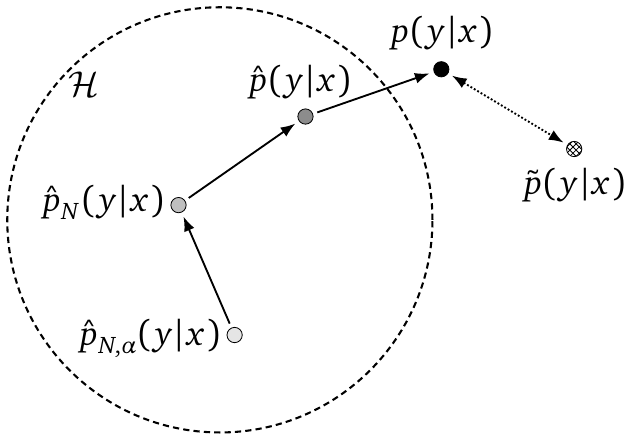}
\caption{Illustration of different sources of epistemic uncertainty in hypothesis space.
Figure taken and adapted from Ref.~\cite{epistemicuncertaintyestimation}.}
\label{fig:hypothesis_space}
\end{figure}

%
\begin{itemize}
\item \textbf{Training variability (algorithmic epistemic uncertainty):}
for fixed data $\mathcal D_N$, variations across $\hat p_{N,\alpha}$ reflect imperfect optimisation or multiple local minima.
A measure is
\begin{equation}
  \Delta_{\mathrm{train}}(N)
  =
  \mathbb E_{\alpha}\!\left[\divf\!\left(\hat p_{N,\alpha},\,\hat p_{N}\right)\right],
\end{equation}
which is reducible by improved optimisation, stabilizing objectives, or ensembling.

\item \textbf{Data variability (finite-sample epistemic uncertainty):}
even under idealised training, $\hat p_N$ depends on the particular dataset.
A measure is
\begin{equation}
  \Delta_{\mathrm{data}}(N)
  =
  \mathbb E_{\mathcal D_N}\!\left[\divf\!\left(\hat p_{N},\,\hat p\right)\right],
\end{equation}
which is reducible by collecting more (informative) data.

\item \textbf{Model bias / misspecification (systematic epistemic uncertainty):}
even with infinite data and perfectly stable training, the best-in-class $\hat p$ may differ from the true $p$ if $\mathcal H$ is misspecified, \ie $p(\cdot| x)\notin\mathcal H$. A conceptual measure is
\begin{equation}
  \Delta_{\mathrm{model}}
  =
  \divf\!\left(p,\,\hat p\right),
\end{equation}
which can be reduced only by enlarging $\mathcal H$, improving inductive biases, or revising modelling assumptions.

\item \textbf{Domain shift / validity uncertainty (out-of-distribution epistemic uncertainty):}
the decomposition above assumes that training data $\mathcal D_N$ and the test data $\tilde{\mathcal D}$ share the same data-generating mechanism.
Denoting the training-domain conditional by $p(y| x)$ and the test-domain conditional by $\tilde p(y| x)$, this corresponds to the assumption $\tilde p(y| x)=p(y| x)$.
If testing inputs or conditions differ (train--test shift, biased simulation, miscalibration), one may have $\tilde p\neq p$ and the learned conditional is queried outside its empirical support. Conceptually, this is epistemic uncertainty about the model's \emph{domain of validity}.
A natural (though typically unobservable) measure is
\begin{equation}
\Delta_{\mathrm{shift}}=\divf\!\left(p,\,\tilde p\right).
\end{equation}
However, whether an uncertainty estimator actually reports increased uncertainty under $\tilde p\neq p$ is method-dependent, since most uncertainty quantification procedures are only calibrated under independent and identically distributed (i.i.d.) assumptions.

\end{itemize}
The first three components above are epistemic because they reflect uncertainty about inferred model objects ($\hat p_{N,\alpha}$, $\hat p_N$, or $\hat p$), rather than irreducible randomness in outcomes.
The fourth component, domain shift/validity uncertainty, is epistemic because it reflects uncertainty about whether the learned conditional is being applied within the domain where the training data are informative.

These epistemic components map differently onto the statistical/systematic axis: training and data variability vanish in the idealized infinite-statistics/ideal-optimisation limit and are therefore \emph{statistical} sources of epistemic uncertainty, while model bias persists unless modelling assumptions are corrected and thus constitutes a \emph{systematic} epistemic uncertainty. Domain shift is likewise \emph{systematic} in practice, as it does not disappear by increasing $N$ in the original domain
and typically requires new data, recalibration, or explicit domain knowledge to resolve.
In contrast, the irreducible width of the data-generating process corresponds to \emph{aleatoric} uncertainty; in probabilistic models it is
represented by the conditional spread of $p(y_\ast| x_\ast,\theta)$ and enters predictions through \cref{eq:posterior_predictive} and the first term in \cref{eq:total_variance_decomp}.

In summary, \emph{statistical/systematic} classify \emph{sources of uncertainty}, whereas \emph{aleatoric/epistemic} classify their
\emph{nature and reducibility}. A consistent uncertainty quantification discussion additionally requires specifying the \emph{object} of uncertainty (inference vs prediction), which we have made explicit through \cref{eq:posterior_predictive,eq:total_variance_decomp}.
Any robust uncertainty quantification framework in AI for physics should make these distinctions explicit and map uncertainties across these dimensions whenever possible.

\subsubsection{Glossary and related terminology}

Beyond the distinctions above, several further terms appear frequently in the literature and can create confusion if not carefully distinguished:

\begin{itemize}

\item \textbf{Predictive distribution / posterior predictive.}
The predictive distribution describes uncertainty over a future outcome $y_\ast$ given input $x_\ast$ and data $\mathcal D$.
In Bayesian language it is given by \cref{eq:posterior_predictive}. 
The term \emph{posterior predictive} emphasizes that uncertainty about model quantities (inference uncertainty) has been marginalized to obtain uncertainty in observable space.

\item \textbf{Statistical and systematic uncertainty in physics. } 
Terminology is not consistent across communities. 
In particle physics, \emph{statistical uncertainty} is sometimes implicitly identified with aleatoric randomness (\eg Poisson counting fluctuations), but strictly speaking it can also include epistemic effects from limited training data or underconstrained parameters. 
Likewise, \emph{systematic uncertainty} in physics often conflates two conceptually different cases: irreducible noise sources such as intrinsic detector fluctuations (aleatoric), and reducible sources such as miscalibration or missing theoretical corrections (epistemic). 
Being explicit about these distinctions is essential for clear communication,especially given that the aleatoric/epistemic categorisation itself depends on the chosen model of analysis~\cite{DerKiureghian2009}.

\item \textbf{Confidence vs.~credibility.}
Frequentist and Bayesian frameworks employ different uncertainty intervals (confidence intervals versus credible intervals). Both are often loosely referred to as \emph{error bars} but their interpretation differs.
A more detailed comparison is provided in \cref{sec:freq_vs_bayesian}.

\item \textbf{Distribution shift / out-of-distribution (OOD).}
These terms refer to situations where the test-time inputs or data-generating mechanism differ from those seen during training.
As discussed above, such shifts can be interpreted as epistemic uncertainty about domain validity, but whether this is detected by an uncertainty
estimator is method-dependent; in physics practice it is often addressed through dedicated validation and robustness tests.

\end{itemize}

\clearpage
\section{Frequentist and Bayesian perspectives on uncertainty}
\label{sec:freq_vs_bayesian}
Uncertainty can be approached from both frequentist and Bayesian perspectives. 
The frequentist view emphasizes long-run guarantees such as coverage of confidence intervals. 
The Bayesian view emphasizes probabilistic modelling of parameters and the posterior predictive distribution. 
The choice between frequentist and Bayesian approaches is not uniform 
across physics communities: particle physics has historically favoured 
frequentist methods -- confidence intervals and hypothesis tests are the 
standard tools for reporting measurements and discovery claims -- while 
astrophysics and cosmology more commonly adopt Bayesian inference for 
parameter estimation and model comparison.
In this section, we outline the key differences and highlight modern tools such as conformal prediction, PAC-Bayes, and Post-Bayesian approaches. 
Throughout, we aim to be explicit about what is held fixed, what is random, and with respect to which probability measure each statement is taken.

\subsection{Frequentist view}

The frequentist view treats parameters $\theta$ as unknown constants, and the observed data $\mathcal D$ as a realization of some random process.
An estimator $\hat \theta = \hat \theta(\mathcal D)$, or for example a confidence interval $[\theta_-(\mathcal D), \theta^+(\mathcal D)]$, is then a random object whose quality, \eg unbiasedness, is assessed in terms of its sampling distribution.
Here, $\theta_-(\mathcal D)$ and $\theta^+(\mathcal D)$ denote the lower and upper endpoints of the interval, both of which are functions of the observed data and therefore random.

\subsubsection*{Confidence intervals and sets}

The classical and central frequentist tool for uncertainty quantification is the confidence interval. An interval ${[\theta_-(\mathcal D), \theta^+(\mathcal D)]}$ is called a $1-\delta$ confidence interval for a parameter $\theta$ if for any~$\theta$
\begin{equation}\label{eq:confidence}
P_{\mathcal D\sim p(\cdot|\theta)}\big(\theta \in [\theta_-(\mathcal D), \theta^+(\mathcal D)]\big) \geq 1 - \delta.
\end{equation}
Here, the probability $P$ is taken over the randomness in $\mathcal{D}$, \ie for repeated drawings of datasets $\mathcal D^{(1)}, \mathcal D^{(2)}, \ldots$, the resulting intervals contain the true value with a fraction at least $1 - \delta$.
Assume for example independently normally distributed observations $x_i \sim \mathcal N(x|\theta,\sigma^2)$. A $1-\delta$ confidence interval for $\theta$ is
\begin{equation}
\left[\hat \theta - z_{1-\delta/2}\frac{\sigma}{\sqrt N}, \hat \theta + z_{1-\delta/2}\frac{\sigma}{\sqrt N}\right]
\qquad \mwith \qquad \hat\theta = \frac{1}{N}\sum_{i=1}^N x_i,
\end{equation}
where $z_{1-\delta/2}$ is the $(1-\delta/2)$-quantile of the standard normal distribution for $N$ observations. For simplicity, we assume $\sigma$ to be known.
A $1-\delta$ \emph{confidence set} $\mathcal C(\mathcal{D})$ is the multidimensional generalization of the interval, \ie $P(\theta \in \mathcal C(\mathcal D)) \geq 1 - \delta$. 
See, \eg Refs.~\cite{casella2024statistical, Wasserman2004} for textbook introductions.
Note that standard confidence intervals guarantee marginal coverage only; stronger notions such as conditional coverage given an ancillary or sufficient statistic are generally not satisfied by common constructions.

\subsubsection*{Conformal prediction}
\label{sec:conformal_prediction}

Intuitively, conformal prediction (CP)~\cite{vovk2005algorithmic,shafer2007tutorialconformalprediction} aims to provide a recipe that yields finite, \ie non-asymptotic, performance guarantees without underlying assumptions about the data distribution or the model, requiring only minor additional assumptions, \eg exchangeability.
CP constructs conformal sets $\mathcal C$ such that 
\begin{equation}
 P(y_\ast \in \mathcal C(x_\ast)) \geq 1 - \delta   
\end{equation}
for target coverage level $1-\delta$.  
In this example, we will stick to the simplest, most common form, \emph{split conformal prediction}. 
Constructing such a conformal set requires five steps:
\begin{enumerate}
    \item Divide the training set $\mathcal D$ into two subsets: a \emph{proper training set} $\mathcal D_\text{train}$ and a \emph{calibration set} $\mathcal D_\text{cal}$ of size $N$. 
    \item Train a model $f$, \eg a neural network, on $\mathcal D_\text{train}$.
    \item Define a measurable \emph{conformal score} $s(x,y) \in \mathbb R$. For regression, this could, \eg be the residual $s(x,y) = |y - f(x)|$, or for a classification model $s(x,y)= 1 - f(x)_y$, \ie one minus the probability of the true class assuming $f(x)$ outputs estimated probabilities. 
    \item Compute $\hat q$ as the $\lceil (N+1)(1-\delta)\rceil/N$-quantile of the scores $s_i = s(x_i,y_i)$ with $i=1,\dots,N$.  
    \item The conformal set for a test point $x_\ast$ is given as $\mathcal C(x_\ast) = \{y: s(x_\ast,y)\leq \hat q\}$. For our regression example, this would be the interval $\mathcal C(x_\ast) = [f(x_\ast) - \hat q, f(x_\ast) + \hat q]$, and for classification, $\mathcal C(x_\ast) = \{y:f(x_\ast)_y\geq 1 - \hat q\}$, where $f(x_\ast)_y$ denotes the estimated probability assigned to class $y$ by the model.
\end{enumerate}
See~\cite{Angelopoulos_2021} for a detailed introduction, including more advanced approaches and Python implementations.

The CP guarantee is \emph{marginal predictive coverage}: the
probability is taken jointly over the randomness in the
calibration data and the test point
$(x_\ast, y_\ast)$, all drawn
exchangeably from the same distribution. In general, CP does
\emph{not} guarantee \emph{conditional} coverage given
$x_\ast$, \ie
\begin{equation}
P(y_\ast \in \mathcal C(x_\ast) \mid x_\ast = x) \geq 1-\delta
\end{equation}
for every $x$. Achieving conditional coverage without
distributional assumptions is impossible in full
generality~\cite{Angelopoulos_2021}.

It is also important to contrast CP with confidence
intervals: CP addresses \emph{predictive uncertainty}
directly as it constructs sets for future observables
$y_\ast$, rather than uncertainty about inferred
parameters $\theta$. This distinction aligns with the
inference-vs-prediction taxonomy of
\cref{sec:taxonomy}.

\subsubsection*{PAC-Bayes}

On a high level, \emph{Probably Approximately Correct (PAC)} bounds~\cite{valiant1984theory} provide guarantees that the error made when choosing a specific model, or hypothesis, is small with high probability. 
Consider a class of models, \eg a parametric class $\mathcal H =\{f_\theta, \theta \in \Theta\}$, \eg $\Theta$ are the weights of a specific neural network architecture. For a given loss function $\loss$ and a data distribution $p$, the unknown \emph{generalization error} or \emph{generalization risk} is given as 
\begin{equation}
    R(\theta) = \mathbb E_{(x,y)\sim p}[\loss(f_\theta(x),y)].
\end{equation}
Given a dataset $\mathcal D =\{(x_i,y_i)\}_{i=1}^N$ drawn from the data distribution $p$, we observe an \emph{empirical risk}
\begin{equation}
    r_{\mathcal D}(\theta) = \frac{1}{N}\sum_{i=1}^N \loss(f_\theta(x_i),y_i),
\end{equation}
with the property that $\mathbb E[r_{\mathcal D}(\theta)] = R(\theta)$. The goal is to find bounds such that
\begin{equation}
    P_{\mathcal D \sim p}\big(R(\theta) \leq r_{\mathcal D}(\theta) + \varepsilon(N,\delta,\mathcal H)\big) \geq 1 - \delta,
\end{equation}
\ie upper bounds on the true generalization error that hold with probability $1 - \delta$. 
The tightness of such a bound depends on the number of observations $N$ (it tightens as $N$ grows), the size of the hypothesis space (a larger set gives looser bounds), and the desired level of confidence (higher confidence requires a looser bound).
As with confidence intervals, the probability here is over the random draw of the dataset $\mathcal D\sim p$, while the hypothesis $\theta$ (or distribution $q$) is held fixed.

\emph{PAC-Bayes}~\cite{mcallester1998some} generalizes this by considering distributions on $\mathcal H$. We will cite the famous McAllester~\cite{mcallester1998some} bound here as a concrete example to discuss the concept.
Under mild technical assumptions, for any $\delta \in (0,1]$ we have
\begin{equation}
P\left(\forall q \text{ on }\mathcal H: R(q) \leq r_{\mathcal D}(q) + \sqrt{\frac{\text{KL}(q\,\|\,p) + \log\frac{2\sqrt{N}}{\delta}}{2N}}\right) \geq 1 - \delta,
\end{equation}
where KL is the Kullback–Leibler divergence~\cite{kldivergence}, $p$ and $q$ are data-independent and data-dependent distributions over $\mathcal H$, with 
\begin{equation}
R(q) = \int_{\Theta} \d q(\theta)\, R(\theta)
\qquad \mand \qquad
r_{\mathcal D}(q) = \int_{\Theta}\d q(\theta)\,r_{\mathcal D}(\theta).
\end{equation}
As $p$ is required to be independent of $\mathcal D$, it is referred to as the prior, and the data-dependent $q$ is known as the posterior.\footnote{The name can mislead: in PAC-Bayes, $p$ and $q$ are not required to be coupled via a likelihood, and the guarantees hold even for ``priors'' $p$ chosen on grounds other than prior belief. Concrete connections to proper Bayesian inference nonetheless exist; for example, minimisation of certain PAC-Bayes bounds recovers Bayesian marginal-likelihood maximisation~\cite{germain2016pac}.}
While the influence of $\mathcal H$ on the tightness of the PAC bound above was merely given by its size, the PAC-Bayesian approach links the tightness of the bound to our prior knowledge over the possible models. In our neural network example, $p$ could, \eg be a standard normal distribution over the weights $\Theta$, while $q$ could be a normal distribution over them with learned means and variances. 
See~\cite{alquier2024user} for a detailed introduction to the topic and for further references.

PAC-Bayes can be viewed as providing a natural bridge between frequentist guarantees and Bayesian-style reasoning: posterior-like distributions arise from optimisation of generalization bounds rather than from Bayes' rule. 
This connection is made more concrete in the discussion of generalized variational inference and Post-Bayesian approaches below.

\subsection{Bayesian view}

From a Bayesian point of view, uncertainty is modelled by placing probability distributions over an unknown parameter $\theta$, and updating this distribution given observed data $\mathcal D$. Bayes' rule combines the prior $p(\theta)$ and the likelihood $p(\mathcal D|\theta)$ into a posterior
\begin{equation}
    p(\theta|\mathcal D)= \frac{p(\mathcal D|\theta)\,p(\theta)}{p(\mathcal D)}
    \qquad \mwith \qquad 
    p(\mathcal D) = \int \d\theta\, p(\mathcal D|\theta)\,p(\theta),
\end{equation}
where the normalizing constant $p(\mathcal D)$ is known as the evidence of the data. 
The posterior $p(\theta|\mathcal D)$ is the primary object of interest in a Bayesian framework. Once it is inferred or approximated, it can subsequently be used for point estimates such as posterior mean and maximum-a-posteriori estimates, for the derivation of credible intervals, or for decision-making and prediction by marginalizing over it rather than having to rely on a single point estimate. 
Assuming well-specified distributions, Bayes' rule is the unique way to coherently update prior beliefs given new data. See~\cite{gelman1995bayesian,mcelreath2018statistical} for textbook introductions.

\subsubsection*{From inference to prediction}

As discussed in \cref{sec:taxonomy},
the posterior $p(\theta|\mathcal{D})$ quantifies
\emph{inference uncertainty}, \ie uncertainty about model
parameters. To obtain \emph{predictive uncertainty} about a future observable $y_\ast$ at input $x_\ast$, one marginalizes over the posterior as in \cref{eq:posterior_predictive}.
%

\subsubsection*{Choosing a prior}

Especially in small sample scenarios, the posterior is sensitive to the chosen prior $p(\theta)$.
When domain knowledge is available, it can be encoded through informative priors, \eg by placing high probability on physically plausible parameter ranges.
In its absence different strategies exist on how to design them.
In \emph{Objective Bayes}~\cite{berger2024objective} priors are designed according to standardized rules to be as ``uninformative'' as possible while maximizing the information to be gained by the data. 
These often lead to improper, \ie unnormalizable, priors but proper posteriors and aim to be default choices unbiased by subjective beliefs.
A second approach is to use \emph{maximum-entropy} priors~\cite{jaynes2003probability}. Given known constraints, \eg support, mean, or variance, the prior is chosen to maximize the Shannon entropy with respect to these constraints, \ie be the least informative but compatible prior (\eg a uniform distribution over a finite support, or a Gaussian if mean and variance are given). 
Finally, \emph{Empirical Bayes}~\cite{efron2012large} is a more pragmatic, data-driven approach. It assumes a parametric prior $p(\theta|\eta)$ and infers the hyperparameters $\eta$ from the data, typically by maximizing the marginal likelihood, \ie the evidence
\begin{equation}
    \hat \eta = \arg\max_\eta \int \d\theta\,p(\mathcal D|\theta)\,p(\theta|\eta).
\end{equation}
This is common, \eg in Gaussian Processes where kernel length-scales or noise variances are often optimized via this evidence maximization, also known as \emph{Type II Maximum Likelihood}.
It is closely related to the evidential learning paradigm where these hyper-parameters $\eta$ are in turn predicted via neural networks.
See \cref{sec:edl} for details, and Ref.~\cite{ulmer2023prior} for a recent introduction on this topic.

\subsubsection*{Credible intervals}

These intervals are the Bayesian analogue to confidence intervals. A $1-\delta$ credible interval on the posterior over $\theta$ is an interval $[\theta_-,\theta^+]$ such that 
\begin{equation}\label{eq:credible}
    P_{\theta \sim p(\theta|\mathcal D)}(\theta \in [\theta_-,\theta^+])\geq 1 - \delta,
\end{equation}
where the probability is with respect to the posterior $p(\theta|\mathcal D)$ for a fixed $\mathcal D$. 
In contrast to confidence intervals, the interval $[\theta_-,\theta^+]$ is fixed for the observed data, while $\theta$ is treated as random under the posterior and lies in the interval with probability at least $1-\delta$. 
Different constructions for these intervals are possible. \emph{Equal-tailed} intervals remove $\delta/2$ of the mass of each tail of the distribution, while \emph{highest posterior density (HPD)} intervals are the smallest interval that contains $1-\delta$ of the posterior mass. See, \eg~\cite{gelman1995bayesian} for a detailed introduction.

\subsubsection*{Credible intervals and frequentist coverage}

Despite the visual similarity between \cref{eq:confidence,eq:credible}, the confidence interval $[\theta_-(\mathcal D),\theta^+(\mathcal D)]$ is random in an aleatoric sense, while $\theta$ is fixed.
In contrast, for \cref{eq:credible} the parameter $\theta$ is random in an epistemic sense, while the data $\mathcal D$ and thus the interval $[\theta_-,\theta^+]$ is fixed. 
As a consequence Bayesian credible intervals do not in general guarantee frequentist coverage and typically only have asymptotic coverage guarantees.
One can nevertheless assess empirical (prior-averaged) coverage via
\begin{equation}
P_{\theta \sim p(\theta),\,\mathcal D \sim p(\mathcal D|\theta)}\big(\theta \in \mathcal{C}(\mathcal D)\big),
\end{equation}
but this is distinct from pointwise (in $\theta$) frequentist coverage.

\subsubsection*{A bridge between frequentist and Bayesian inference}

The Bernstein--von Mises (BvM) theorem links frequentist and Bayesian inference. Roughly speaking, under suitable regularity conditions and well-specified models, the inferred posterior distribution $p(\theta |\mathcal D)$ becomes asymptotically normal and concentrates around the maximum likelihood estimator $\theta_\text{MLE}$; that is, it converges in distribution as
\begin{equation}\label{eq:bvm}
    \sqrt{N}(\theta - \hat{\theta})\xrightarrow{\text{dist}}  \mathcal N(0, I(\theta_0)^{-1}),
\end{equation}
where $\theta_0$ is the unknown true parameter and $I(\theta_0)$ is the Fisher information at $\theta_0$.
Thus, asymptotically, the influence of the prior vanishes, and Bayesian credible intervals/sets coincide with their frequentist confidence intervals/sets, despite their interpretations remaining distinct, as discussed above.
In finite samples, however, conditional coverage
discrepancies between credible and confidence sets may
persist, particularly when the prior is informative or the model is misspecified.
See Ref.~\cite{van2000asymptotic} for further details.

\begin{figure}
\centering
\includegraphics[width=0.95\linewidth]{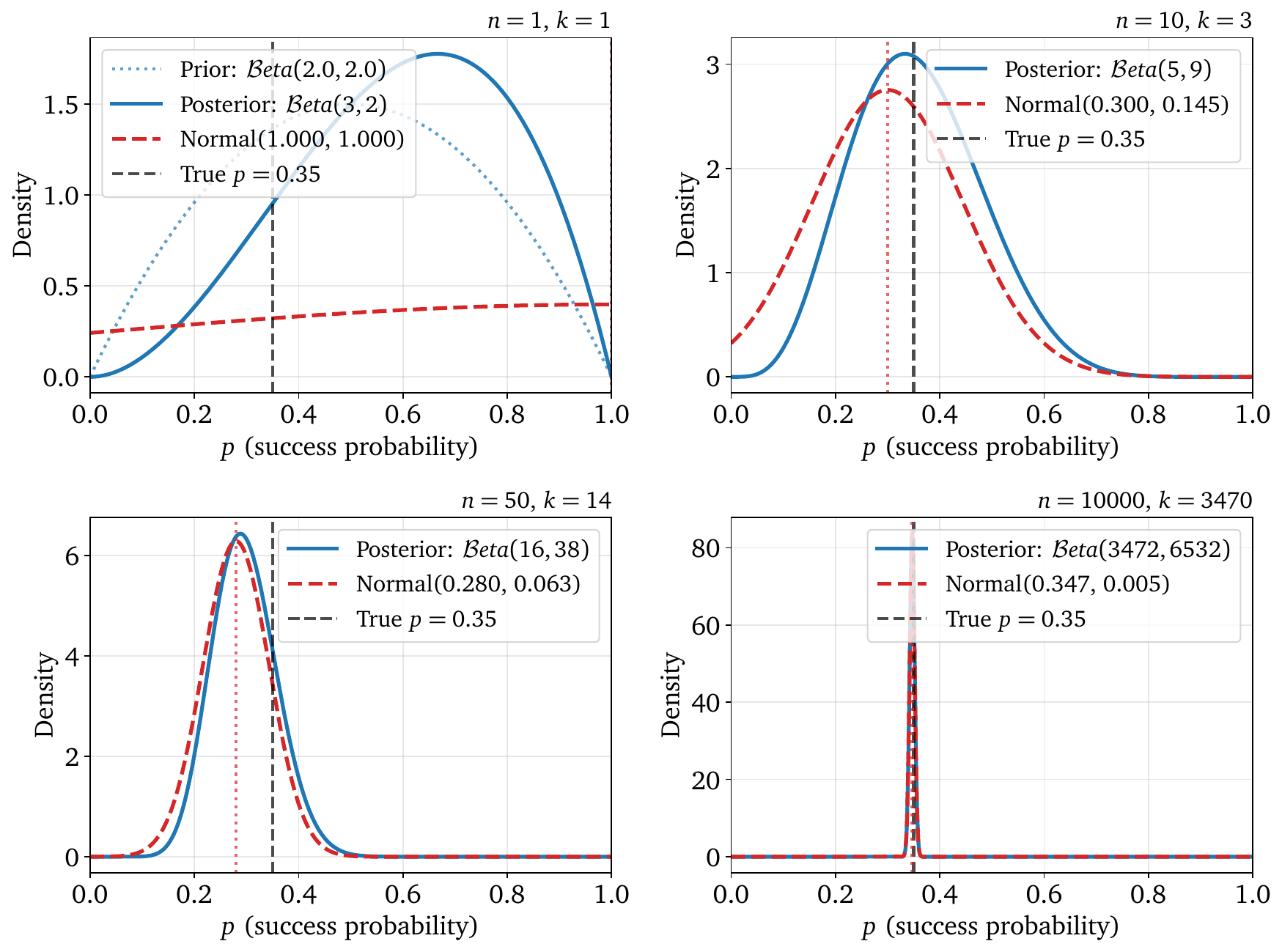}
\caption{\emph{Bernstein-von Mises convergence visualization.} 
We assume observations coming from a Bernoulli distribution over $\{0,1\}$ with $p = P(X = 1) = 0.35$ (vertical black dashed line in each plot), and a Beta prior over $p$, denoted as $\Betadist(2,2)$, whose density is shown in the upper-left plot.
For $N \in \{1,10,50,\num{10000}\}$ samples, with $k$ instances where class $1$ was sampled, we visualize
the analytical posterior density, ${\Betadist(2 + k, 2 + N-k)}$, together with the maximum likelihood estimate (red dotted vertical line), and the density of a normal distribution centred at the MLE with variance given by the inverse Fisher information. 
For $N=1$ the posterior remains close to the prior 
as the single observation provides little information.
As given by BvM (see \cref{eq:bvm}), the two coincide as $N \to \infty$.
}
\label{fig:bvm_beta}
\end{figure}

See \cref{fig:bvm_beta} for a concrete example.
Consider observations drawn from a Bernoulli distribution with unknown $p=0.35$, with $N$ denoting the total number of observations and $k$ the number of observations of class~1.
We construct two estimators. The frequentist maximum likelihood estimator is the fraction of observations of class 1, corresponding to $\hat{\theta}$ in \cref{eq:bvm}, \ie $\hat{\theta} = k/N$. The Bayesian estimator is based on the posterior distribution. Starting from a Beta prior, $\Betadist(2,2)$, conjugate to the Bernoulli distribution, the posterior is analytically tractable as $\Betadist(2+k,2+N-k)$, where $N$ is the total number of observations and $k$ is the number of ones.
Plotting both the posterior and $\mathcal N(\hat{\theta}, I(\theta_0)^{-1})$ shows that they coincide as the number of observations increases and concentrate on the true probability $p$.

\subsubsection*{Inferring a posterior}

In practice, calculation of the posterior $p(\theta|\mathcal D)$ is usually not analytically tractable, and approximation techniques are required.
\emph{Sampling-based Monte Carlo (MC)} approaches approximate the posterior via a set of samples $\{\theta_1,\ldots, \theta_S\}$, without requiring access to its normalizing constant. 
Classical Markov chain Monte Carlo (MCMC) algorithms, such as Metropolis-Hastings or Hamiltonian Monte Carlo (HMC) (see~\cite{betancourt2017conceptual} for an introduction) are asymptotically exact, but are often computationally expensive, especially in high-dimensional settings and large amounts of data.
In scenarios with large datasets, stochastic-gradient-based methods such as SGHMC~\cite{chen2014stochastic} have been proposed.
\emph{Variational Inference (VI)}~\cite{blei2017variational}, in contrast, approaches posterior inference as an optimisation problem, by learning a variational posterior $\hat q$ that approximates the intractable posterior such that
\begin{equation}
 \hat q=\arg\min_{q\in \mathcal Q}\text{KL}(q(\theta)\,\|\,p(\theta|\mathcal D)),   
\end{equation}
for a fixed family of distributions $\mathcal Q$. 
This is equivalent to choosing $q$ as the distribution that maximizes the \emph{Evidence lower bound (ELBO)} given as
\begin{equation}
    \mathbb E_q[\log p(\mathcal D|\theta)] - \text{KL}(q(\theta)\,\|\,p(\theta))\leq \log p(\mathcal D),
\end{equation}
which solely relies on tractable distributions.
If the chosen family is parametric, \ie $\mathcal Q = \{q_\psi:\psi \in \Psi\}$, this objective can be optimized via standard stochastic gradient descent approaches. However, its scalability and computational tractability comes at the cost of providing a biased solution, as $p(\theta|\mathcal D)\not\in \mathcal Q$ in all relevant scenarios.
Finally, a third classical approach (originally introduced by Laplace already back in 1774~\cite{laplace1774memoire}) is the \emph{Laplace approximation (LA)}, which approximates the log-posterior locally by a quadratic Taylor-expansion around the mode $\hat \theta$, such that 
\begin{equation}
p(\theta|\mathcal D) \approx \mathcal N(\theta|\hat\theta,\Sigma), \text{ where } \Sigma = (-\nabla_\theta^2\log p(\theta|\mathcal D)|_{\hat \theta})^{-1}.
\end{equation}
These approximations are simple and often effective, but similar to variational inference approaches inherently local and thus biased. See~\cite{mackay1992bayesian,daxberger2021laplace} on successful applications of LA in Bayesian deep learning.
Since none of these approximations recover the true posterior $p(\theta|\mathcal D)$, the resulting credible intervals may be miscalibrated even asymptotically, which motivates the empirical coverage and calibration tests discussed in \cref{sec:evaluation}.

\subsubsection*{Generalized and Post-Bayesian approaches}

Bayes' rule links the prior and posterior via a likelihood $p(\mathcal D|\theta)$.
While theoretically attractive, it suffers from several limitations that become relevant, especially in modern large-scale or deep-learning-based problems and approaches. Its assumptions of well-specified priors and likelihoods, and especially computational feasibility, are difficult to ensure in practice. A growing research field known as \emph{Generalized Bayes}, or \emph{Post-Bayes}, is therefore focused on weakening these restrictions while keeping the philosophical approach intact. We give a short overview of one of them, \emph{Generalized Variational Inference}~\cite{knoblauch2022an}, and refer the reader to a recent series of lectures~\cite{postbayes_seminar} that provides a deep introduction to the research field and its current results.
Generalized variational inference switches from inferring the posterior to an optimisation-centric approach, similar to what is done in variational inference~\cite{blei2017variational}, by solving the following objective,
\begin{equation}
    \hat q = \arg\min_{q\in \mathcal Q}\left\{\loss(q,\mathcal D) + \divf(q,p_0)\right\},
\end{equation}
where $\mathcal{D}$ is the data, $\loss$ is an arbitrary loss function, \ie the term concerned with fitting a model to the dataset, $q\in \mathcal Q$ is a distribution from a family of distributions/hypotheses $\mathcal Q$, and $\divf(\cdot,\cdot)$ is a divergence measure between two distributions that regularizes $q$ to be close to a prior distribution $p_0$. All four, $\mathcal Q$, $\loss$, $\divf$, and $p_0$, can be chosen by the practitioner independently of each other. Standard variational inference is recovered, \eg by letting $\loss$ be the negative log-likelihood, $\divf$ the Kullback-Leibler divergence, and $\mathcal Q$ and $p_0$ be the variational posterior and prior.

The modularity of GVI and similar approaches opens connections to several
perspectives beyond standard Bayesian inference. For
instance, certain choices of loss and divergence recover
objectives related to the PAC-Bayes generalization
bounds discussed above, while others connect to
robustness under model misspecification or to
decision-theoretic formulations. The key insight is that by
decoupling the components of the inference objective, GVI
allows practitioners to tailor the statistical properties of
the resulting posterior-like distribution to the requirements
of their specific application.

\clearpage
\section{Validation and diagnostics}
\label{sec:evaluation}

Having established a glossary for the various sources of uncertainty and discussed their statistical interpretation, we now turn to commonly used methods to \emph{validate} uncertainty estimates. 
The ordering is important: diagnostics such as coverage, calibration, and bias are only meaningful once it is clear what is treated as random, what object the uncertainty refers to, and with respect to which probability measure a given statement is made. 
Throughout this section, our primary focus is on \emph{predictive} uncertainty, \ie uncertainty about future or unobserved outcomes rather than uncertainty about model parameters themselves.

The \emph{true} data-generating conditional distribution $p_{\rm truth}=p(y|x)$ is never observed directly, and the learning algorithm searches within a restricted hypothesis class ${\cal H} = \{p_{\theta}(y|x)\}$, \eg neural networks with a fixed architecture. Given a dataset $\mathcal{D}_N$ of size $N$, and an optimisation procedure with parameter $\alpha$ -- indicating algorithmic randomness such as initialization, stochastic optimisation, mini-batch ordering, or early stopping -- training produces the predictive conditional distribution $\hat p_{N,\alpha}(y|x)$. 
From $\hat{p}_{N,\alpha}(y|x)$ one typically derives predictive regions, for example prediction intervals 
\begin{equation}
    \mathcal{C}_{N,\alpha}(x) \subseteq Y,
\end{equation}
or more generally any functional of the predictive distribution used for decision-making or downstream inference. 
Unlike the confidence and credible intervals discussed in \cref{sec:freq_vs_bayesian}, which quantify \emph{inference uncertainty} about model parameters or hypotheses, the set $\mathcal{C}_{N,\alpha}(x)$ is a \emph{predictive} region in observable space: it is meant to contain a future outcome $y$ at input $x$. 

Uncertainty validation relies on several complementary statistical approaches.
Coverage and calibration tests, discussed in \cref{subsec:coverage}, assess whether predictive uncertainty statements are statistically reliable. 
Bias tests, discussed in \cref{subsec:bias}, probe whether central estimates or derived quantities are systematically shifted away from their target. 
The two tests are complementary. Bias tests probe the accuracy of central predictions or derived quantities, 
while coverage tests assess the reliability of the associated predictive uncertainty estimates. 
A given method may exhibit correct coverage while being biased, or be unbiased but poorly calibrated. 
Robust uncertainty validation requires that both properties be simultaneously satisfied. In physics applications, where ML-based predictions often enter quantitative analyses and precision measurements, the joint control of bias and coverage is particularly critical. 
Proper scoring rules, discussed in \cref{subsec:score}, provide a more global assessment of the full predictive distribution.
Together, the diagnostics that we will briefly discuss in this section provide a principled framework for assessing whether probabilistic ML models can be reliably integrated into scientific workflows.

\subsection{Coverage tests}
\label{subsec:coverage}

A statistical \emph{coverage test} assesses whether the random predictive region produced by our learning 
procedure contains the true future outcome $y_\ast$ with the advertised frequency, namely whether the empirical coverage of prediction intervals matches their nominal coverage. 
Formally, for a nominal level $1-\delta$, coverage requires
\begin{equation}
   P_{N,\alpha,(x_\ast,y_\ast)\sim p_{\rm truth}} \bigl[ y_\ast \in \mathcal{C}_{N,\alpha}(x_\ast)\bigr] = 1 - \delta . 
\end{equation}
Here the predictive region $\mathcal{C}_{N,\alpha}(x_\ast)$ is itself a random object, since it is constructed from the random training dataset $\mathcal{D}_N$ and the algorithmic randomness $\alpha$; the probability is therefore taken jointly over these sources of randomness as well as over a fresh test point $(x_\ast,y_\ast)$.
This is therefore a statement about \emph{marginal predictive coverage}. 
It does not imply conditional coverage at fixed input $x_\ast$, nor does it test inference-level coverage of model parameters. 
The distinction is important, since conditional predictive coverage is in general a much stronger requirement and is typically unattainable without further assumptions.

We note that a coverage test does not tell us whether the model is correct. 
Even if the hypothesis class ${\cal H}$ is misspecified, \ie if $p_{\rm truth}$ lies outside ${\cal H}$, a method can still have correct coverage. 
What is tested is whether the pipeline
\begin{equation}
(\mathcal{D}_N,\alpha)\longrightarrow \hat{p}_{N,\alpha}(y|x) \longrightarrow  \mathcal{C}_{N,\alpha}(x)    
\end{equation}
correctly accounts for all sources of uncertainty that affect predictions. 
For instance, if one constructs $95\%$ predictive intervals, approximately $95\%$ of true values should fall within these intervals when evaluated over repeated draws of training and test data. 
Systematic deviations between empirical and nominal coverage indicate some kind of miscalibration, either over-confident uncertainty estimates (when the empirical coverage is smaller than the nominal coverage) or under-confident estimates (when the empirical coverage is larger than the nominal one).

A closely related notion is \emph{calibration}. 
A well-calibrated model should, when predicting an event with probability $p$, observe that event occurring with frequency $p$ in the long run. 
The Expected Calibration Error (ECE)~\cite{pmlr-v70-guo17a} quantifies this notion by partitioning predictions into $M$ bins indexed by $m=1,\ldots,M$, and measuring the average absolute difference between predicted confidence and empirical accuracy within each bin, namely
\begin{equation}
    \mathrm{ECE} = 
\sum_{m=1}^M \frac{|B_m|}{N}
\bigl|\mathrm{acc}(B_m) - \mathrm{conf}(B_m) \bigr|,
\end{equation}
where $B_m$ is the set of samples whose predicted confidence falls into bin $m$, and the empirical accuracy in bin $m$ is given by
\begin{equation}
\mathrm{acc}(B_m)=\frac{1}{|B_m|}\sum_{i \in B_m}\mathbf{1}\{\hat y_i = y_i\},
\end{equation}
with the true $y_i$ and the predicted label $\hat{y}_i$ in classification problems $\in\{0,1\}$, 
and the average predicted confidence in bin $m$ is given by
\begin{equation}
\mathrm{conf}(B_m)
= \frac{1}{|B_m|} \sum_{i \in B_m} \hat p_i.
\end{equation}
The disadvantage of ECE is that it is sensitive to binning choices and can mask subtle calibration failures. 
Ref.~\cite{pmlr-v70-guo17a} shows that modern neural networks, despite high accuracy, often exhibit poor calibration, with ECE providing a practical diagnostic for this phenomenon. 
Alternative calibration diagnostics include reliability diagrams, which visualise the same information graphically, and kernel-based calibration tests~\cite{kernelcalib2019,kernelcalib2022} that avoid binning entirely.

Coverage and calibration are related but distinct notions of reliability. 
Coverage is a property of \emph{predictive regions}: a set 
$\mathcal{C}_{1-\delta}(x)$ has correct coverage if it contains the true outcome $y_\ast$ with frequency $1-\delta$ over repeated draws of training data and test points. 
It is a binary check -- in or out -- at a specified nominal level $\delta$. 
Calibration is a property of \emph{predicted probabilities}: a model is calibrated if, whenever it assigns probability $p$ to an event, that event occurs with frequency $p$ in the long run. 
Calibration therefore concerns the reliability of 
the full probability output across all levels simultaneously, not just whether a single interval covers the truth. 
A model can have correct coverage at one nominal level while being miscalibrated at others; conversely, a well-calibrated model does not automatically produce correctly sized predictive regions unless those regions are constructed 
consistently from its probability outputs. 
Both properties are necessary for trustworthy uncertainty quantification.

\subsection{Bias tests}
\label{subsec:bias}

Statistical \emph{bias} tests complement coverage analysis by examining whether predictions or derived quantities are systematically shifted away from their target. 
In practice, such shifts are often diagnosed through normalised residuals or pull distributions, which compare deviations from the truth to the quoted uncertainties.
Under correct uncertainty quantification, Gaussian model assumptions, and in the large-sample limit, normalised residuals should follow a standard normal distribution.
Under these assumptions, deviations from this expectation -- a non-zero mean, a width different from unity, or distortions in shape -- can reveal problems in the uncertainty quantification pipeline, although heavy tails or asymmetries may also simply reflect a genuinely non-Gaussian predictive distribution.

The key point is that bias is always defined relative to a target, and that target must be specified.
If the target is the true conditional distribution $p_{\rm truth}$, then bias concerns the expected estimator
\begin{equation}
    \mathbb{E}_{\mathcal{D}_N,\alpha}\!\left[\hat p_{N,\alpha}(y |x)\right]
\end{equation}
and compares it to $p_{\rm truth}(y |x)$, when the data is generated from the true data-generating distribution itself.  
The test is very effective in determining systematic errors in the learning procedure, such as model misspecification or optimisation bias (early stopping, local minima). 
Even with infinite data, bias can persist if ${\cal H}$ cannot represent $p_{\rm truth}$ or if the hyperparameters of the learning algorithm are not adequate.

In practice, however, bias is usually assessed not for
the full conditional distribution, but for a functional
$T\!\left(p(\cdot|x)\right)$ of interest, such as the mean, a quantile, a class probability, a likelihood ratio, or other physically relevant observable. Denoting
\begin{equation}
    \tau(x)=T\!\left(p_{\rm truth}(\cdot|x)\right),
    \qquad
    \hat\tau_{N,\alpha}(x)=T\!\left(\hat p_{N,\alpha}(\cdot|x)\right),
\end{equation}
a bias test examines whether
\begin{equation}
\mathbb{E}_{\mathcal{D}_N,\alpha}\!\left[\hat\tau_{N,\alpha}(x)\right] = \tau(x)
\end{equation}
holds, at least within statistical precision. Bias tests therefore probe whether the learning procedure is systematically shifted away from the target
when averaged over repeated datasets and training randomness. In contrast to coverage, which directly tests the reliability of predictive regions, bias
focuses on systematic displacement of central quantities or other derived functionals.

In the context of particle physics, typical bias tests are often operationalized through \emph{multi-closure tests}, in which one generates multiple 
synthetic datasets from a known reference model and repeats the full inference or learning procedure on each of them. This makes frequentist notions such as 
repeated-sample bias and dispersion practically testable even when only one real dataset is
available. Concrete examples can be found in
Refs.~\cite{Barontini:2025lnl,Bahl:2025xvx,Harland-Lang:2024kvt,DelDebbio:2021whr,Costantini:2025agd}.
More concretely, let $F$\footnote{In some cases $F$ is a simple function or an integral transform of the underlying law convolved with known operators.} denote a forward map from latent quantities to observable central values,
\begin{equation}
   \hat y_i = F(x_i),
\end{equation}
and let synthetic replicas be generated as
\begin{equation}
   y_i^{(k)} = \hat y_i + \eta_i^{(k)}
   \qquad \mwith \qquad
   \eta^{(k)}\sim\mathcal N(0,\Sigma),
\end{equation}
where $\Sigma$\footnote{The covariance matrix is 
sometimes provided by experimentalists and sometimes modelled or estimated in the inference process} is the covariance matrix specifying uncertainties and correlations. Each replica $k=1,\dots,K$ represents one synthetic realization of the
experiment. Fitting the inference model to each replica yields an ensemble of predictions or estimators 
\begin{equation}
\tilde{y}^{(k)}\sim\hat{p}_{N,\alpha}(y^{(k)}|x),
\end{equation}
that can then be compared to the known truth.
If the target functional $T$ is simply the predicted observable $y$ itself, a common diagnostic is the normalised bias estimator
\begin{equation}
    \label{eq:bias_definition}
    B^{(k)}
    =
    \frac{1}{N}\sum_{i,j=1}^{N}
    \left(\tilde y_i^{(k)}-\hat y_i\right)
    (\tilde \Sigma^{-1})_{ij}
    \left(\tilde y_j^{(k)}-\hat y_j\right),
\end{equation}
where $\tilde \Sigma$ is either the experimental covariance matrix or the covariance estimated by the model itself. Related is the \emph{pull}
quantity
\begin{equation}
\label{eq:pull_definition}
t_i
=
\frac{\mathbb{E}_k\left[\tilde y_i^{(k)}\right] - y_i^{\rm true}}
{\sigma_i} \qquad \mwith \qquad
\sigma_i=\sqrt{\tilde \Sigma_{ii}},
\end{equation}
which compares the average deviation from the truth to the quoted uncertainty.
Under well-calibrated Gaussian uncertainties, the pull distribution should be approximately centred at zero with unit variance.
Averaging \cref{eq:bias_definition} over the ensemble of replicas yields
\begin{equation}
  \label{eq:bias_variance_ratio_definition}
    R_b=\mathbb{E}_k\left[ B^{(k)}\right].
\end{equation}
Under faithful Gaussian uncertainty estimates, each term in 
\cref{eq:bias_definition} follows a $\chi^2$ distribution with $N$ degrees 
of freedom, so that $\mathbb{E}_k[B^{(k)}] = 1$ and therefore $R_b \simeq 1$.
Values $R_b<1$ indicate that uncertainties are overestimated (the 
covariance $\tilde\Sigma$ is too large relative to the observed residuals), 
while $R_b>1$ indicates underestimated uncertainty or excess residual 
dispersion beyond what $\tilde\Sigma$ accounts for.

\subsection{Scoring rules}
\label{subsec:score}

To conclude, it is worth mentioning \emph{proper scoring rules}, which provide a complementary and more holistic approach to uncertainty validation by directly assessing the quality of the full predictive distribution rather than only selected summaries such as intervals or central values. 
A scoring rule assigns a numerical score to a predictive distribution and an observed outcome, and is called \emph{proper} if its expected value is optimised when the predicted distribution coincides with the true data-generating distribution~\cite{Gneiting01032007}. 
As a result, proper scoring rules incentivise honest and well-calibrated probabilistic predictions, discouraging artificial inflation or deflation of uncertainties.

From this perspective, coverage and bias tests can be viewed as partial diagnostics of distributional quality. Coverage probes the reliability of
predictive regions, while bias tests examine systematic shifts in selected functionals. 
Proper scoring rules subsume both aspects by simultaneously penalising miscalibration, excessive dispersion, and systematic deviations in location or shape.

For categorical outcomes $y_i \in \{1,\ldots,C\}$, a standard example is the Brier score~\cite{1950brier}, a strictly proper scoring rule that measures the mean squared deviation between predicted class probabilities and observed outcomes:
\begin{equation}
\mathrm{BS}_C = \frac{1}{N} \sum_{i=1}^N \sum_{c=1}^C
\left(\hat{p}(y=c|x_i) - \mathbf{1}\{y_i = c\}\right)^2,
\end{equation}
where $\mathbf{1}\{y_i = c\}$ equals one if observation $i$ belongs to class $c$ and zero otherwise.
Perfect predictions yield a score of $\mathrm{BS}=0$. For binary classification ($C=2$), this reduces to 
\begin{equation}
\mathrm{BS}_2 = \frac{1}{N}\sum_{i=1}^N (\hat p_i - y_i)^2  \qquad \mwith \qquad \hat p_i = \hat p(y=1|x_i).
\end{equation}
For continuous outcomes, the negative log-likelihood is a strictly proper scoring rule. In the Bayesian framework, this corresponds to evaluating the log-score of the posterior predictive distribution defined in \cref{eq:posterior_predictive}, which explicitly marginalizes over inference uncertainty $p(\theta|{\mathcal{D}})$ and therefore rewards models that assign high probability mass to observed data while penalising overconfident or miscalibrated predictions.
Aggregated measures such as the \emph{log pointwise predictive density} extend this idea to full datasets.

Proper scoring rules formalise the fundamental trade-off between \emph{calibration} and \emph{sharpness}~\cite{Gneiting01032007}. 
While perfectly calibrated predictions can be obtained by issuing overly diffuse distributions, such forecasts lack practical utility. 
Conversely, overly sharp predictions that neglect uncertainty lead to poor scores when observations 
fall in low-probability regions. 
By construction, proper scoring rules reward concentration of probability mass subject to statistical consistency, thereby providing a principled balance between informativeness and reliability.

In this sense, scoring rules offer a unified criterion for uncertainty validation: while coverage and bias tests diagnose specific failure modes, proper scoring rules evaluate the overall fidelity of probabilistic predictions. Together, these approaches form a complementary toolkit for assessing whether learned predictive distributions can be trusted in scientific inference.

\clearpage
\section{Quantification methods and examples}
\label{sec:examples_methods}

In this section, we illustrate and compare different approaches to uncertainty quantification (UQ) on simple, controlled examples. 
We focus on supervised learning tasks, in particular regression and classification, where predictive uncertainties can be visualised directly in terms of predictive intervals and assessed using standard diagnostics such as coverage, calibration, and pull distributions. 
While these examples are deliberately low-dimensional, the underlying principles are generic: epistemic versus aleatoric uncertainty, the role of inductive biases, and the impact of model assumptions on extrapolation and out-of-distribution behaviour.
The same concepts extend naturally to generative modelling, where the goal is no longer to predict a conditional mean but to learn an entire probability distribution. 
In that case, UQ manifests itself in the fidelity of the learned density, in uncertainty on derived observables, and in the robustness of sampling and reweighting procedures. 
Throughout, our aim is not to advocate a single ``best'' method, but to use simple benchmarks to expose how different modelling assumptions translate into different uncertainty estimates, and how these can be validated with statistically well-defined tests.

\subsection{Methods}
\label{subsec:methods}

We describe five UQ strategies which reflect different modelling 
assumptions, four of which are illustrated in the worked examples below.
Gaussian processes (GPs) provide a non-parametric Bayesian reference with an explicit func\-tion-space prior.
Bayesian neural networks (BNNs) place a prior on network parameters and
approximate the posterior using, for instance, variational inference (VI) or Hamiltonian Monte Carlo (HMC).
Repulsive ensembles (REs) approximate a posterior sample with a finite set of diverse models by augmenting training with an explicit diversity regulariser.
Evidential deep learning (EDL) estimates uncertainties in a single 
forward pass without sampling or ensembling; it is included here for 
completeness as it has found application in physics, but is not used in the worked examples below.
Finally, conformal prediction constructs finite-sample predictive sets with distribution-free marginal coverage guarantees under exchangeability.

\subsubsection{Bayesian neural networks}
\label{sec:bnn}

Bayesian neural networks (BNNs) extend deterministic networks by treating the model as a random object and placing a prior either on the network parameters or, more abstractly, directly in function space. In the parameter-space view one assigns a prior $p(\theta)$~\cite{mackay1992bayesian} and combines it with the likelihood $p(\mathcal{D}| \theta)$ to form the posterior $p(\theta| \mathcal{D})\propto p(\mathcal{D}|\theta)\,p(\theta)$. 
For BNNs, predictive uncertainty is obtained by marginalizing the network likelihood over this posterior, as in \cref{eq:posterior_predictive}.
Alternatively, one may specify a prior over functions $p(f)$~\cite{sun2018functional,rudner2022tractable}, which induces a distribution over models directly; in practice, function-space priors are often realized implicitly through architectural choices together with a parameter prior. Gaussian processes (GPs), as discussed in more detail later, are the canonical example for explicit function-space constructions and certain infinite-width limits of neural networks recover GP priors.

The posterior $p(\theta| \mathcal{D})$ is typically intractable for deep networks, so approximate inference is required. 
A range of strategies exist, broadly divided into sampling-based methods (MCMC, HMC) that are asymptotically exact but computationally expensive, and optimisation-based methods (variational inference, Laplace approximation) that are more scalable but introduce systematic bias due to the restricted approximating family.
We briefly summarize variational inference and HMC here, as the former is used in the regression example below and HMC in the classification example.

In variational inference (VI) one approximates the posterior by a tractable family $q_\phi(\theta)$ by maximizing the evidence lower bound (equivalently minimizing the negative ELBO),
\begin{equation}
\loss_{\rm BNN}(\phi)
= \kl\!\bigl(q_\phi(\theta)\,\|\,p(\theta)\bigr)
-\mathbb{E}_{\theta\sim q_\phi}\!\bigl[\log p(\mathcal{D}| \theta)\bigr].
\label{eq:loss_bnn_elbo}
\end{equation}
For heteroscedastic regression with $p(y| x,\theta)=\mathcal N(\mu_\theta(x),\sigma^2_\theta(x))$, this yields both a learned mean and a learned uncertainty proxy.
Given samples $\theta_k\sim q_\phi$, the predictive mean and variance at $x$ are estimated as
\begin{align}
\hat\mu(x) &\equiv \frac{1}{K}\sum_{k=1}^K \mu_{\theta_k}(x), \notag\\
\hat\sigma^2_{\rm tot}(x) 
&\equiv \underbrace{\frac{1}{K}\sum_{k=1}^K\!\big(\mu_{\theta_k}(x)-\hat\mu(x)\big)^2}_{\text{epistemic}}
+\underbrace{\frac{1}{K}\sum_{k=1}^K \sigma^2_{\theta_k}(x)}_{\text{aleatoric}}.
\label{eq:bnn_epi_alea}
\end{align}
Other posterior approximations are possible, including sampling-based MCMC methods~\cite{chen2014stochastic}, Laplace approximations~\cite{daxberger2021laplace}, and MC Dropout~\cite{gal2016dropout}, but VI provides a scalable default for deep models.
BNNs have found application across several particle physics, cosmology, and astrophysics tasks, including jet and particle tagging~\cite{Bollweg:2019skg,Kasieczka:2020vlh}, topo-cluster energy calibration~\cite{ATLAS:2024rpl}, amplitude surrogates~\cite{Badger:2022hwf,Bahl:2024gyt,Bahl:2025xvx}, generative modelling~\cite{Bellagente:2021yyh,Heimel:2023mvw,Huetsch:2024quz} and generative amplification~\cite{Bieringer:2024nbc}, event reconstruction~\cite{Fanelli:2023lmp}, cosmological redshift estimates~\cite{Drabicki:2026rgn,Soriano:2026mwu}, and astrophysical source classification~\cite{Butter:2021mwl}.
See Ref.~\cite{Papamarkou2024} for a broader discussion and further references.

An alternative to variational approximations is \emph{Hamiltonian Monte Carlo (HMC)}~\cite{betancourt2017conceptual}, a gradient-based MCMC method that exploits the geometry of the posterior landscape.
HMC augments the parameter vector $\theta$ with an auxiliary momentum variable $r$ and simulates Hamiltonian dynamics on the joint energy $H(\theta,r)=-\log p(\theta|\mathcal{D})+\tfrac{1}{2}r^\top r$ using leapfrog integration, followed by a Metropolis acceptance step.
By following the curvature of $\log p(\theta|\mathcal{D})$, HMC proposes distant yet high-probability states, reducing the random-walk behaviour that limits simpler MCMC schemes such as Metropolis--Hastings.
Given sufficient samples and a well-tuned integrator, HMC is asymptotically exact and therefore provides a principled reference for the true posterior, against which cheaper approximations can be benchmarked.
Its main limitation is computational cost: each leapfrog trajectory requires multiple gradient evaluations over the full dataset, which becomes prohibitive for large-scale problems.
Stochastic-gradient variants such as SGHMC~\cite{chen2014stochastic} mitigate this by sub-sampling data at each step, at the price of introducing a bias whose magnitude depends on the noise correction scheme employed.
In the classification example of \cref{sec:classification}, we use HMC as a ``gold standard'' posterior sampler and compare its uncertainty estimates against those obtained from VI and ensemble methods.
Predictive moments are estimated as in \cref{eq:bnn_epi_alea}, with the variational samples $\theta_k\sim q_\phi$ replaced by HMC draws $\theta_k\sim p(\theta|\mathcal{D})$.

\subsubsection{Evidential deep learning}
\label{sec:edl}

Evidential deep learning (EDL)~\cite{sensoy2018evidentialclass, amini2019deepregression, ulmer2023prior} estimates aleatoric and 
epistemic uncertainties in a single forward pass, without ensembling or sampling. 
Rather than placing a prior over the network weights as in BNNs, EDL promotes the likelihood parameters $\lambda$ themselves to random variables and places a conjugate prior over them, such that marginalising over $\lambda$ yields a predictive distribution in closed form whose negative log-likelihood becomes the objective to be minimised. 
The choice of conjugate prior depends on the task.
\begin{description}
\item[Evidential regression:] 
For regression, one typically starts from a Gaussian likelihood,
\begin{align}
    p(y|\lambda) = \mathcal{N}(y|\,\bar{y}, \sigma^2),
\end{align}
and the conjugate prior over $\lambda = (\bar{y}, \sigma^2)$ is the 
Normal-Inverse-Gamma (NIG) distribution,
\begin{align}
    p(\lambda|m) = \frac{\beta^\alpha \sqrt{v}}{\Gamma(\alpha)\sqrt{2\pi\sigma^2}}
    \left(\frac{1}{\sigma^2}\right)^{\alpha+1}
    \exp\!\left(-\frac{2\beta + v(\gamma - \bar{y})^2}{2\sigma^2}\right),
\end{align}
with evidential parameters $m = \{\gamma, v, \alpha, \beta\}$, where 
$\gamma$ is the predicted mean, $v, \alpha, \beta > 0$ control the 
spread of the prior, and all four are output directly by the network, i.e., we have a dependence on $x$, $m(x)=\{\gamma(x),\nu(x),\alpha(x),\beta(x)\}$.
Marginalising over $\lambda$ yields a Student-$t$ predictive distribution in closed form,
\begin{align}
    p(y|x,m) = \mathrm{St}\!\left(y\,\Big|\,\gamma(x),\,\frac{\beta(x)(1+v(x))}{v(x)\alpha(x)},\,2\alpha(x)\right), \qquad \Phi(x) = 2v(x) + \alpha(x),
\end{align}
where the total evidence $\Phi$ encodes the strength of belief in the predicted parameters; higher $\Phi$ corresponds to lower epistemic uncertainty.
We suppress the dependence of $m$ on the features $x$ in the remaining equations of this section for notational convenience.
The predicted mean and uncertainties follow analytically as
\begin{align}
    \hat{y} = \gamma, \qqquad
    \sigma^2_\text{alea} = \frac{\beta}{\alpha - 1}, \qqquad
    \sigma^2_\text{epi}  = \frac{\beta}{v(\alpha - 1)}.
    \label{eq:edl_uncertainties_reg}
\end{align}
The network is trained by minimising the negative log Student-$t$ likelihood together with an evidence regulariser~\cite{amini2019deepregression,meinert2022multivariate},
\begin{align}
    \loss_\text{ER} = - \sum_i \log\left( \text{St}\left(y\bigg|\gamma_i, \frac{\beta_i(1+v_i)}{v_i\alpha_i}, 2\alpha_i\right)\right) + \lambda_R \sum_i 
    |y_\text{train}(x_i) - \gamma_i|\cdot\Phi_i,
    \label{eq:loss_er}
\end{align}
where $\lambda_R>0$ is a tunable hyperparameter.

\item[Evidential classification:]
For classification with $K$ classes, one starts from a categorical likelihood 
\begin{equation}
  p(y|x,\lambda) = \mathrm{Cat}(y|P)  
\end{equation}
over class probabilities $P = (P_1,\ldots,P_K)$, and the conjugate prior is the Dirichlet distribution~\cite{sensoy2018evidentialclass,malinin2018predictive},
\begin{align}
    p(P|m) = \mathrm{Dir}(P|\alpha) 
    = \frac{\Gamma(\Phi)}{\prod_{k=1}^K \Gamma(\alpha_k)}
    \prod_{k=1}^K P_k^{\alpha_k - 1},
    \qquad \Phi = \sum_{k=1}^K \alpha_k,
\end{align}
where the concentration parameters $\alpha = (\alpha_1,\ldots,\alpha_K)$ with 
$\alpha_k > 0$ are output directly by the network and $\Phi$ is the total evidence.
Marginalising over $P$ yields a Dirichlet-categorical predictive 
distribution, from which the predicted class probabilities and per-class variance follow from the moments of the Dirichlet,
\begin{align}
    \hat{P}_k = \frac{\alpha_k}{\Phi}, \qqquad
    \mathrm{Var}(P_k) = \frac{\alpha_k(\Phi - \alpha_k)}{\Phi^2(\Phi+1)},
\end{align}
where $\mathrm{Var}(P_k)$ captures total uncertainty in the predicted 
class probability and decreases as $\Phi$ grows.
Unlike the regression case, the Dirichlet does not yield a clean per-class aleatoric/epistemic split; a scalar summary of epistemic uncertainty is the vacuity~\cite{sensoy2018evidentialclass} $u = K/\Phi$, which equals unity when no evidence is present ($\alpha_k = 1\ \forall k$, \ie $\Phi = K$) and vanishes as total evidence accumulates.
The network is trained by minimising the Bayes risk under the L2 loss 
together with a KL regulariser~\cite{sensoy2018evidentialclass},
\begin{align}
    \loss_\text{EDL} = \sum_i \sum_{k=1}^K \left[
    \left(\hat{P}_{ik} - y_{ik}\right)^2 
    + \frac{\hat{P}_{ik}(1 - \hat{P}_{ik})}{\Phi_i + 1}
    \right]
    + \lambda_t\,\mathrm{KL}\!\left[
    \mathrm{Dir}(\tilde{\alpha}_i)\,\|\,\mathrm{Dir}(1)
    \right],
    \label{eq:loss_edl}
\end{align}
where $\tilde{\alpha}_{i} = y_i + (1 - y_i)\,\alpha_i$ 
are the evidence-adjusted concentrations after removing non-misleading evidence for the correct class, and $\lambda_t$ is an annealing coefficient that gradually increases the regularisation effect over training.
\end{description}
Both methods require only a single forward pass at inference time, but their interpretation as faithful epistemic uncertainty estimators has been called into question: depending on the loss formulation and regulariser, the learned ``epistemic'' component does not always behave as a posterior variance, and may not vanish in the infinite-data limit~\cite{Meinert_2023,juergens2024}.
EDL has been applied in particle physics to amplitude surrogates~\cite{Bahl:2025xvx}, jet identification~\cite{Khot:2025kqg}, and BSM model discrimination~\cite{Kriesten:2024ist}, though we do not use it in the worked examples below.

\subsubsection{Repulsive ensembles}

Deep ensembles~\cite{lakshminarayanan2017} approximate epistemic uncertainty by training $K$ models with different initialisations.
However, naive ensembles can underestimate uncertainty if members collapse to similar functions.
Repulsive ensembles~\cite{DBLP:journals/corr/abs-2106-11642} introduce an interaction between ensemble members inspired by particle-based variational inference to cover a broader set of plausible solutions.
Denoting by $f_\theta(x)$ the model output and by $\widehat f$ a stop-gradient, we write the (stochastic) objective evaluated on mini-batches as
\begin{equation}
\loss_{\rm RE}
= \frac{1}{K}\sum_{i=1}^{K}
\left[
-\frac{1}{B}\sum_{b=1}^B\log p(y_b| x_b,\theta_i)
+ \frac{\beta}{N}\,
\frac{\sum_{j=1}^K \mathcal{K}\bigl(f_{\theta_i},\widehat f_{\theta_j}\bigr)}{\sum_{j=1}^K \mathcal{K}\bigl(\widehat f_{\theta_i},\widehat f_{\theta_j}\bigr)}
+\frac{1}{N}\log p(\theta_i)\right],
\label{eq:loss_re_abstract}
\end{equation}
where $\mathcal K$ is a positive-definite kernel (typically an RBF kernel) evaluated on mini-batches, and $N$ denotes the full training set size, ensuring that the repulsion and prior contributions are normalised with respect to the full dataset.
In the idealized derivation in which the ensemble members can be interpreted as particles approximating a true posterior, one should set $\beta=1$; in practice, $\beta$ can be tuned to regulate training dynamics and the degree of diversity.
As for BNNs, total predictive variance is estimated from ensemble spread plus predicted noise.

Standard deep ensembles~\cite{lakshminarayanan2017}, where diversity arises from random initialisation rather than an explicit repulsion term, are among the most widely used practical methods for epistemic uncertainty estimation due to their simplicity and strong empirical performance.
A prominent physics example of ensemble-based UQ is the NNPDF replica 
method~\cite{Forte:2002fg,Ball:2008by,NNPDF:2021njg,Kassabov:2022orn,Costantini:2024wby,Barontini:2025lnl}, in which an ensemble of neural networks is trained on bootstrapped data replicas; the spread across replicas provides an uncertainty estimate on the parton distribution functions that propagates faithfully through to collider predictions.
Repulsive ensembles extend this idea by introducing an explicit interaction between members to encourage broader coverage of the solution space, and have been applied in particle physics to amplitude surrogates~\cite{Bahl:2024gyt,Bahl:2025xvx} and topo-cluster energy calibration~\cite{ATLAS:2024rpl}, as well as in cosmology for parameter inference~\cite{Rover:2024pvr}.

\subsubsection{Gaussian processes}

A Gaussian process (GP) specifies a prior over functions $f$ via a mean function $m(\cdot)$ and kernel $k(\cdot,\cdot)$,
such that for any inputs $x$ the function values are jointly Gaussian~\cite{rasmussen2006gaussian}.
For regression with Gaussian observation noise $\sigma^2$, the predictive distribution at a test point $x_\ast$ is Gaussian,
\begin{align}
p\!\left(f_\ast| x_\ast, \mathcal D\right)
&=\mathcal N\!\left(\mu_\ast,\,\sigma_\ast^2\right), \notag \\[1em]
\mwith \qquad
\mu_\ast = k_\ast^\top (K+\sigma^2 I)^{-1} y\,,
&\qqquad
\sigma_\ast^2 = k(x_\ast,x_\ast) - k_\ast^\top (K+\sigma^2 I)^{-1} k_\ast,
\label{eq:gp_predictive}
\end{align}
where $K_{ij}=k(x_i,x_j)$ and $(k_\ast)_i=k(x_i,x_\ast)$. Uncertainty outside the training domain is therefore controlled explicitly by the kernel choice. 
The distribution above is the posterior over the latent function value $f_\ast$.
For a noisy test observation $y_\ast = f_\ast + \varepsilon$ with 
$\varepsilon\sim\mathcal N(0,\sigma^2)$, the predictive distribution acquires 
an additional noise term,
\begin{equation}
p\!\left(y_\ast| x_\ast, \mathcal D\right)
= \mathcal N\!\left(\mu_\ast,\,\sigma_\ast^2 + \sigma^2\right).
\end{equation}
In the regression example of \cref{subsec:regression} we use the 
noisy predictive distribution for interval construction.
Gaussian processes have already found several applications in particle physics, for example in Bayesian approaches to inverse problems and PDF closure studies, PDF determination, and the reconstruction of QCD spectral functions from lattice data~\cite{Frate:2017mai,Horak:2021syv,Horak:2023xfb,Candido:2024hjt,Medrano:2025cmg}.

\subsubsection{Conformal prediction}

Conformal prediction provides distribution-free prediction sets with finite-sample marginal coverage under exchangeability, by calibrating nonconformity scores on a held-out calibration set.
In this section we use conformal prediction to construct $1-\delta$ intervals around a base regressor; details and assumptions are summarized in \cref{sec:conformal_prediction}.
While the coverage guarantee is model-agnostic, the resulting interval shape and efficiency inherit the inductive bias of the underlying model.
In particular, conformal intervals guarantee \emph{marginal} coverage over the joint randomness of calibration and test data drawn exchangeably from the same distribution; they do not guarantee conditional coverage at a fixed input $x_\ast$, nor do they control extrapolation behaviour outside the support of the calibration data.
Conformal prediction has been applied in particle physics~\cite{Araz:2025vuw}, cosmology~\cite{Leterme:2026cqj}, and in astrophysics, including gravitational-wave analyses~\cite{Malz:2024zjd, Ashton:2024wae}, radio galaxy classification~\cite{walls2026montecarloconformalprediction}, and solar flare regression~\cite{hong2026uncertaintyawaresolarflareregression}.

\subsection{Regression}
\label{subsec:regression}

\begin{figure}[t!]
    \centering
    \includegraphics[width=0.495\textwidth]{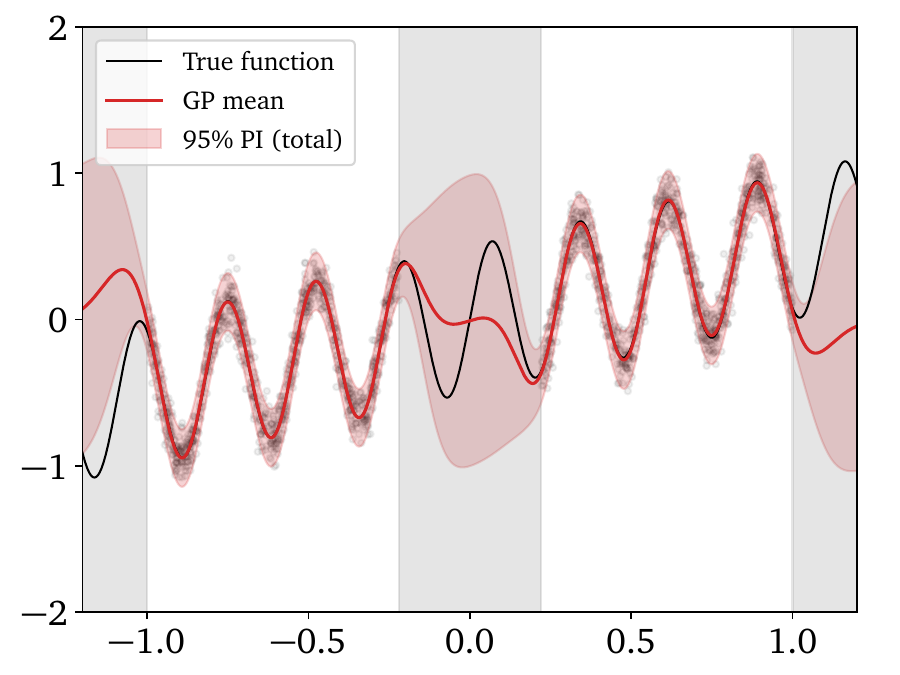}
    \includegraphics[width=0.495\textwidth]{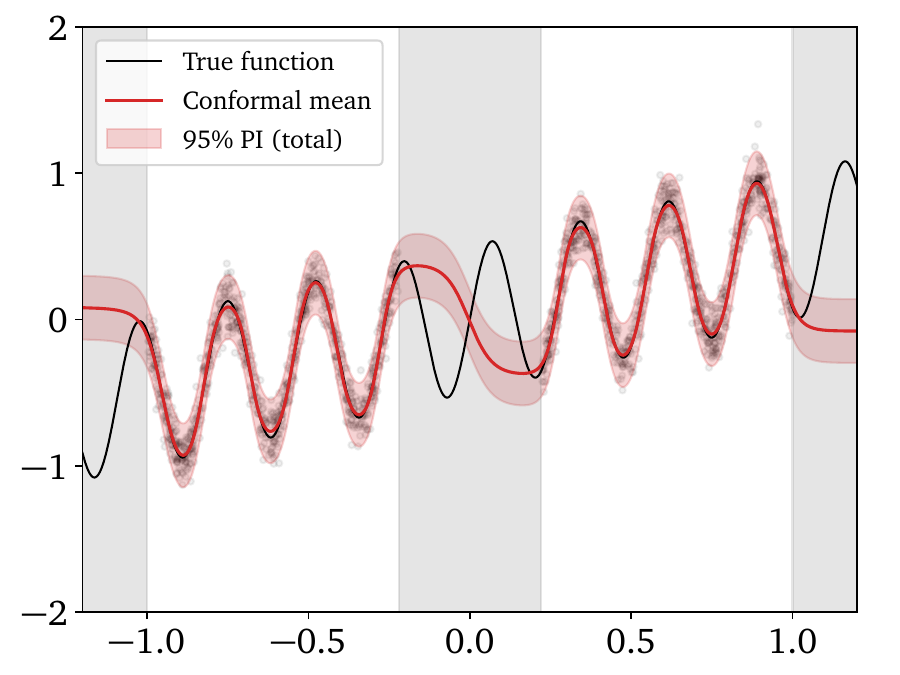} \\
    \includegraphics[width=0.495\textwidth]{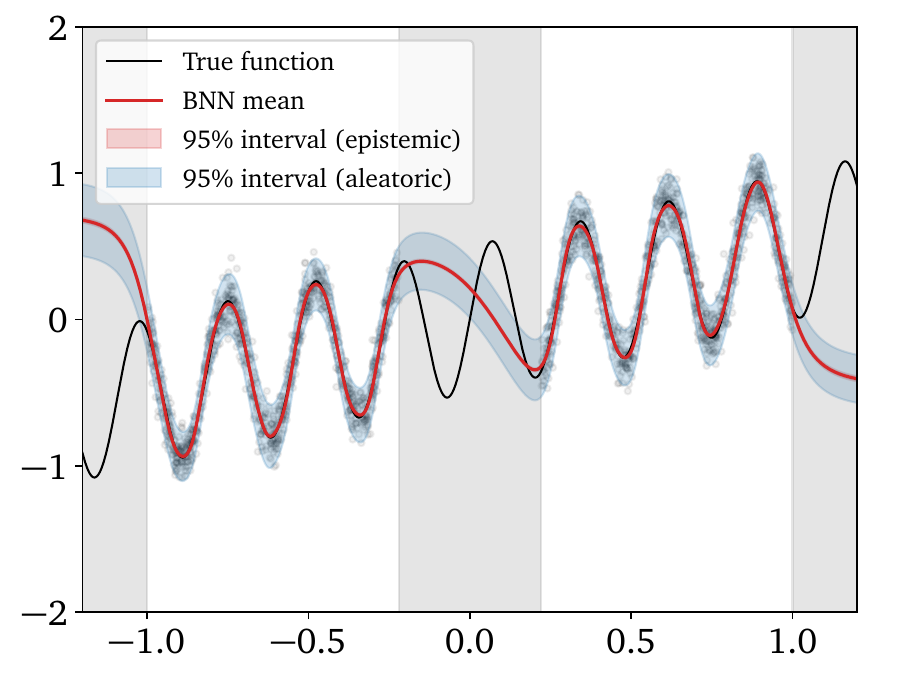}
    \includegraphics[width=0.495\textwidth]{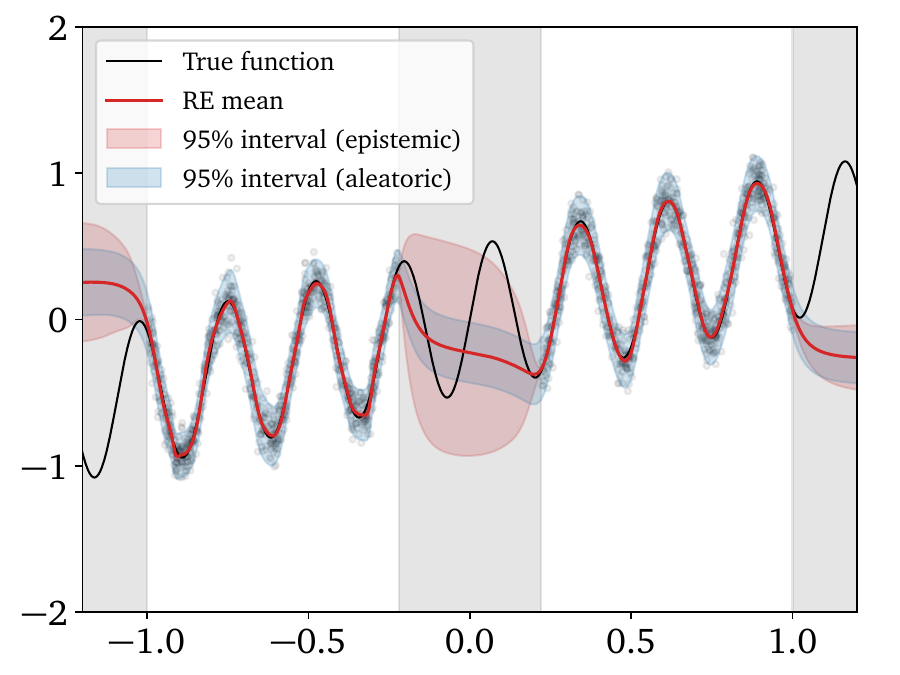}
    \caption{Toy regression example comparing four uncertainty quantification methods: Gaussian Process (top left), neural network with conformal prediction (top right), Bayesian neural network (bottom left), and repulsive ensemble (bottom right). The black curve denotes the true target function, the red curve the predictive mean, and the shaded bands indicate $95\%$ predictive intervals (PI). For the Bayesian neural network and repulsive ensemble, the total predictive uncertainty can additionally be decomposed into epistemic and aleatoric components. Gray vertical regions indicate extrapolation beyond the training domain.}
    \label{fig:regression_toy}
\end{figure}

As a first illustration, we consider a one-dimensional toy regression problem, shown in \cref{fig:regression_toy}. 
Training data are sampled only in a finite interval, while the target function is also evaluated outside this region, allowing us to probe both interpolation and extrapolation.

We compare the four methods introduced above: a Gaussian process (GP), conformal prediction applied to a neural-network regressor, a Bayesian neural network (BNN) in the mean-field approximation using variational inference, and a repulsive ensemble (RE). 
In all panels, the black curve denotes the true target function, the red curve the predictive mean, and the shaded bands indicate $95\%$ predictive intervals (PI).
For the BNN and RE, the total uncertainty can furthermore be decomposed into epistemic and aleatoric contributions as in \cref{eq:bnn_epi_alea}.

Inside the training region, all methods reproduce the target function with comparable accuracy and similar uncertainty bands. 
In this regime, the data is sufficiently informative and differences in prior assumptions and model class have only limited impact on the predictive distribution.
The situation changes in the extrapolation regions. 
There, the data no longer constrains the solution directly, and the behaviour of the predictive mean and uncertainty bands is governed by the respective inductive biases of the different approaches. 
For the GP this is set explicitly by the kernel, while for the neural-network-based methods it is encoded implicitly through the architecture, training objective, and posterior approximation. 
Conformal prediction inherits its interval shape from the underlying regressor and calibration sample, but does not by itself prescribe how the model should behave outside the training domain.

This example highlights a central lesson of UQ: in poorly constrained regions, uncertainty estimates are not determined by the data alone, but also by the modelling assumptions used to extend beyond them. 
Methods that appear similarly successful in interpolation can therefore lead to visibly different uncertainty estimates in extrapolation and out-of-distribution settings.

\subsubsection{Validation diagnostics}
\label{subsubsec:regression_validation}

\begin{figure}[t!]
    \centering
    \includegraphics[width=0.495\textwidth]{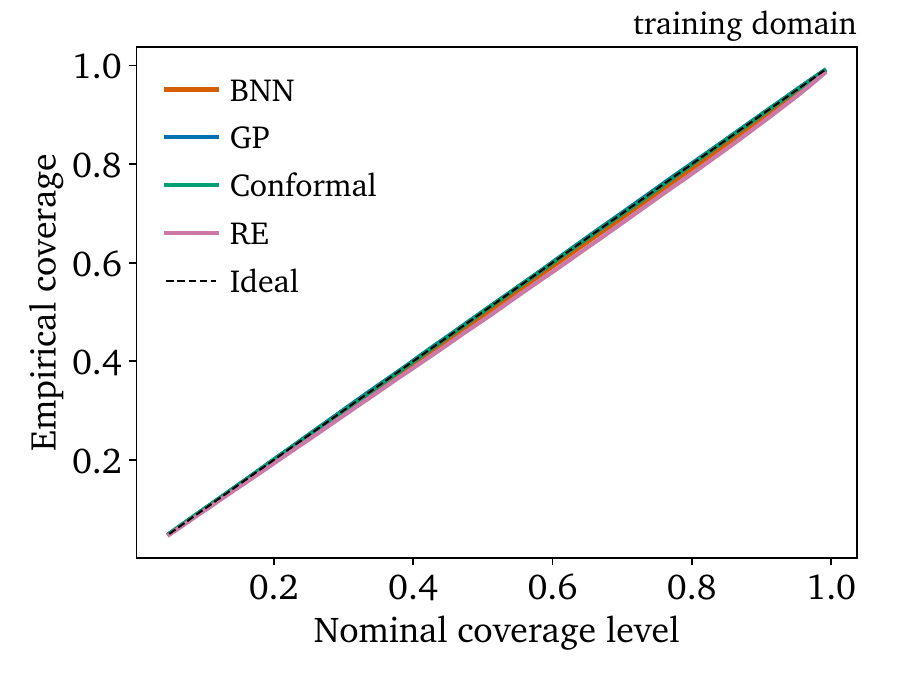}
    \includegraphics[width=0.495\textwidth]{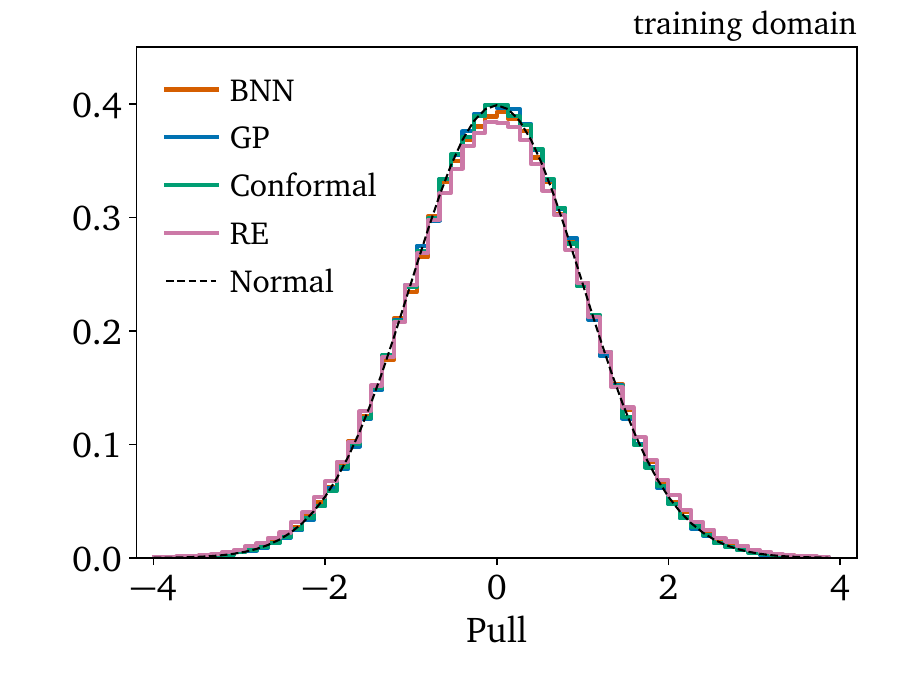}\\
    \includegraphics[width=0.495\textwidth]{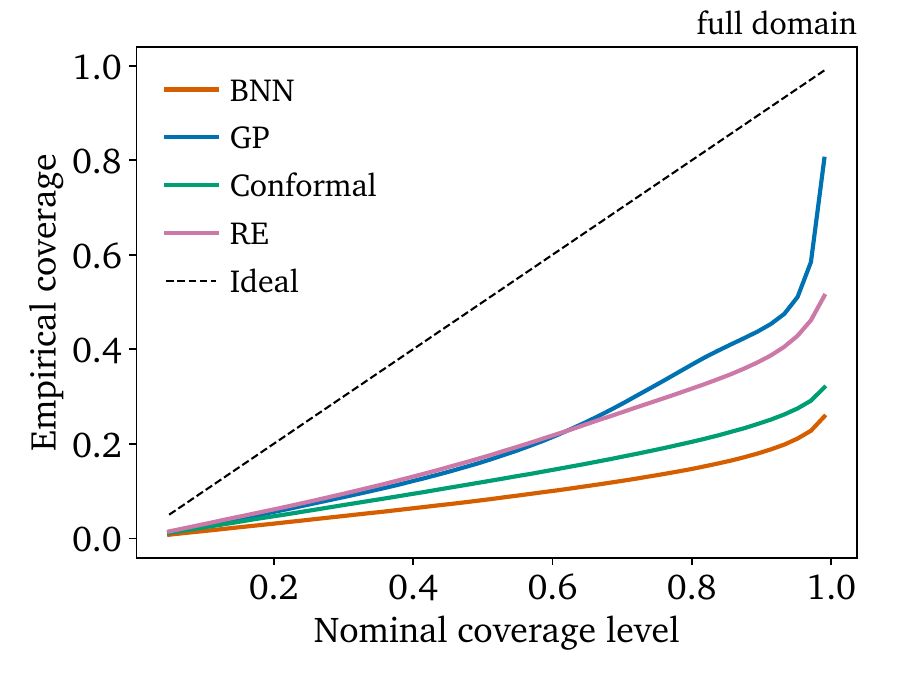}
    \includegraphics[width=0.495\textwidth]{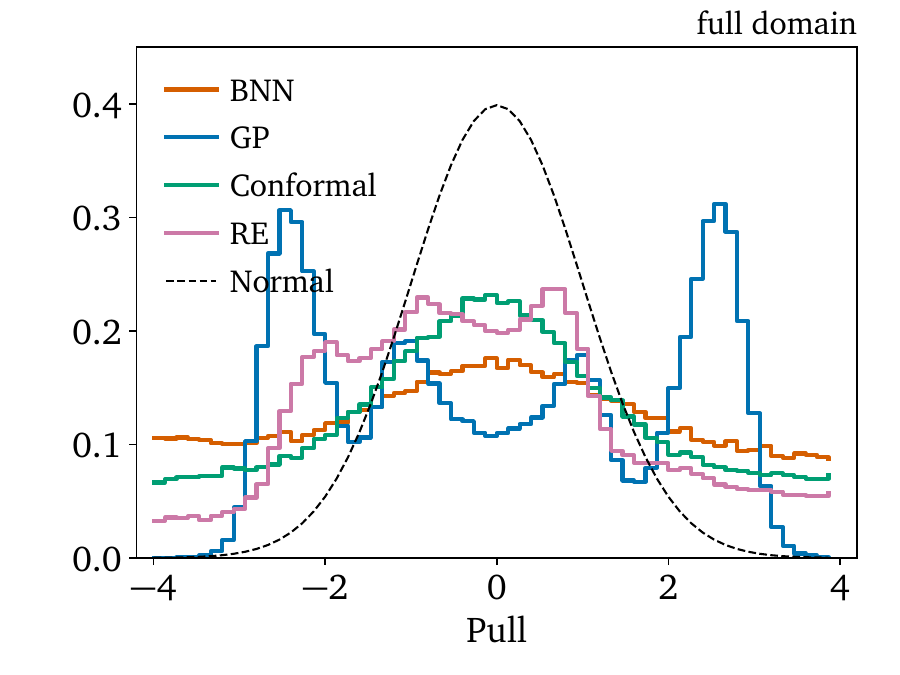}
    \caption{Validation diagnostics for the toy regression example. Upper row: diagnostics evaluated on test points inside the training domain only. Lower row: the same diagnostics evaluated on the full test domain, including extrapolation points outside the training support. Left column: calibration curves for interval coverage, comparing marginal empirical coverage to the nominal coverage level; the diagonal corresponds to perfect calibration. Right column: pull distributions defined in \cref{eq:pull_definition}; a well-calibrated probabilistic model is expected to yield pulls centred around zero with unit width.}
    \label{fig:regression_validation}
\end{figure}

To assess the statistical quality of the quoted uncertainties, we complement the qualitative comparison in \cref{fig:regression_toy} with validation diagnostics shown in \cref{fig:regression_validation}. 
These tests probe whether the predictive intervals are numerically consistent with the observed residuals and therefore provide a more stringent notion of calibration than visual agreement alone.
To keep the present toy example simple, we restrict ourselves to two representative diagnostics, namely a calibration curve comparing empirical and nominal interval coverage, and pull distributions. In realistic applications, however, uncertainty validation would generally need to be broader and also include diagnostics such as bias tests and proper scoring rules, as discussed in \cref{sec:evaluation}.

The left column of \cref{fig:regression_validation} shows calibration curves, comparing marginal empirical coverage to the nominal interval level.
For each nominal level $1-\delta$, we compute the fraction of test targets contained in the corresponding prediction interval.
A perfectly calibrated method would lie on the diagonal, while curves below (above) the diagonal indicate undercoverage (overcoverage), corresponding to intervals that are systematically too narrow (too wide).
In the upper-left panel, where the evaluation is restricted to the training domain, all methods achieve near-nominal coverage.
In the lower-left panel, where extrapolation points are included, this picture breaks down: all methods show clear deviations from the diagonal, 
indicating that nominal interval levels no longer translate into reliable frequentist coverage once predictions are made beyond the support of the training data.

The right column of \cref{fig:regression_validation} shows the pull distributions, using the definition in \cref{eq:pull_definition}, evaluated on an independent test sample.
For methods with an explicit predictive standard deviation, a well-calibrated and approximately Gaussian predictive distribution should yield pulls centred at zero with unit width and close to a standard normal shape.
This behaviour is observed in the upper-right panel, where only test points from the training region are considered, indicating that the quoted uncertainties are consistent with the observed residual fluctuations in interpolation.
By contrast, in the lower-right panel, where the full domain including extrapolation is used, the pull distributions broaden and distort for all methods.
This shows that none of the approaches considered here provides fully reliable uncertainty estimates once the model is forced into extrapolation, where the predictions are dominated by modelling assumptions rather than direct statistical constraint.

Taken together, these diagnostics confirm the qualitative picture of \cref{fig:regression_toy}: interpolation is comparatively insensitive to the precise UQ framework, whereas extrapolation exposes both the conceptual differences between methods and the general failure of their uncertainty estimates to remain well calibrated outside the training support.

\subsection{Classification}
\label{sec:classification}

As a second illustration, we turn to a simple binary classification problem. 
Compared to regression, classification highlights different aspects of uncertainty quantification: predictive uncertainty is naturally expressed in terms of class probabilities and entropy-based diagnostics, and the distinction between aleatoric and epistemic uncertainty appears through class overlap versus lack of training support.

Our example is based on the classical two-moons dataset from \texttt{Scikit-learn}~\cite{scikit-learn}, where the center square is removed from the training data.\footnote{While we restrict ourselves in this discussion to a two-class classification scenario, everything generalizes to multi-class classification.}
Our baseline is a multi-layer perceptron (MLP) with three layers and a width of 32 per layer.
We extend it with four approaches to perform UQ: 
\begin{itemize}
    \item An ensemble of $M$ MLPs trained independently of each other.
    \item An ensemble of $M$ MLPs trained jointly as a repulsive ensemble, as introduced above.
    \item A BNN version of the architecture trained with mean-field variational inference.
    \item Hamiltonian Monte Carlo (HMC) as our ``gold standard'' for inferring the posterior.
\end{itemize}
The regularization on the weights of each model is scaled to correspond to a standard normal prior, independent of whether the models are deterministic, and we use $M=10$.

Regression allows for a natural decomposition of predictive uncertainty into aleatoric and epistemic contributions via the law of total variance.
In classification, however, predictive uncertainty is more naturally characterized using information-theoretic~\cite{cover2006elements} quantities.
Given a dataset $\mathcal D$, the posterior predictive distribution at input $x$ is
\begin{equation}
    p(y|x,\mathcal D)=\mathbb E_{\theta\sim p(\theta|\mathcal D)}[p(y|x,\theta)].
\end{equation}
The \emph{total predictive uncertainty} at $x$ is measured by the entropy of the posterior predictive, defined as
\begin{equation}
    \text{ent}(p(y|x,\mathcal D)) = -\sum_{y} p(y|x,\mathcal D)\log p(y|x,\mathcal D),
\end{equation}
which for binary classification reduces to the entropy of a Bernoulli random variable. 
A key property is that this entropy decomposes as
\begin{equation}
    \text{ent}(p(y|x,\mathcal D)) = \mathbb E_{\theta \sim p(\theta|\mathcal D)}[\text{ent}(p(y|x,\theta))] + \text{MI}(y,\theta|x,\mathcal D),
\end{equation}
where $\text{MI}(y,\theta|x,\mathcal D)$ denotes the mutual information~\cite{MacKay2003,cover2006elements} between the prediction $y$ and the parameters $\theta$, conditioned on input $x$ and dataset $\mathcal D$.

The first term is the expected predictive entropy of an individual model sampled from the posterior. 
It is commonly interpreted as \emph{aleatoric uncertainty}, \ie the intrinsic uncertainty that remains even if the true model were known. 
We note, however, that this interpretation is exact only when individual models are perfectly calibrated: a miscalibrated model (such as a BNN with a restrictive mean-field approximation) may inflate this term with unresolved epistemic uncertainty, causing the decomposition to underestimate the mutual information term~\cite{depeweg2018decomposition}.
The second term captures \emph{epistemic uncertainty}. It quantifies how much different models disagree in their predictions and allows for two equivalent interpretations.
First, we have 
\begin{equation}
    \text{MI}(y,\theta|x,\mathcal D) = \mathbb E_{y\sim p(y|x,\mathcal D)}[\text{KL}(p(\theta|y,x,\mathcal D)\,\|\,p(\theta|\mathcal D))],
\end{equation}
\ie the expected reduction in posterior uncertainty after observing $y$ at input $x$. This interpretation is primarily used in active learning~\cite{houlsby2011bayesian}.
The second is that 
\begin{equation}
    \text{MI}(y,\theta|x,\mathcal D) = \mathbb E_{\theta \sim p(\theta|\mathcal D)}[\text{KL}(p(y|x,\theta)\,\|\,p(y|x,\mathcal D))],
\end{equation}
\ie the expected deviation of an individual model’s predictions from the posterior predictive.
See, e.g.,~\cite{depeweg2018decomposition,chen2025bayesian} for further details.
In practice, the true posterior $p(\theta|\mathcal D)$ is intractable, and it is necessary to rely on Monte Carlo samples $\{\theta_m\}_{m=1}^M$.
These allow approximation of each of these quantities as
\begin{align}
    p(y|x,\mathcal D) &\approx \hat p(y|x,\mathcal D)= \frac1M\sum_m p(y|x,\theta_m),\notag\\
    \text{ent}(p(y|x,\mathcal D))&\approx \text{ent}(\hat p(y|x,\mathcal D)),\notag\\
    \mathbb E_{\theta\sim p(\theta|\mathcal D)}[\text{ent}(p(y|x,\theta))] &\approx \frac1M\sum_m \text{ent}(p(y|x,\theta_m)),
\end{align}
and finally $\text{MI}(y,\theta|x,\mathcal D)$ as the difference between the last two. 
In our concrete example, these samples, $\{\theta_m\}_{m=1}^M$, are (i) the parameters of an ensemble of $M$ members, (ii) samples from a variational approximation to the true posterior, ${q(\theta)\approx p(\theta|\mathcal D)}$, or (iii) samples from an HMC sampler which, assuming convergence, are samples from the true posterior. 

\begin{figure}
    \centering
    \includegraphics[width=0.45\linewidth]{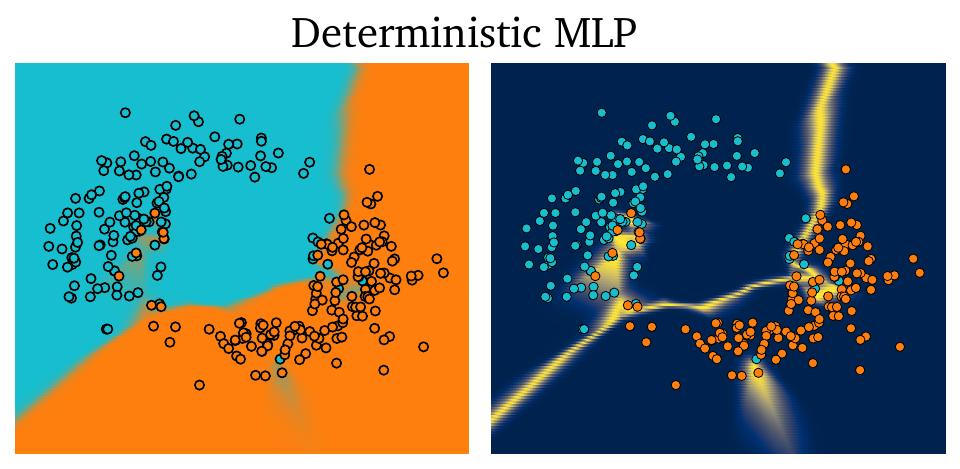}\\
    \includegraphics[width=0.98\linewidth]{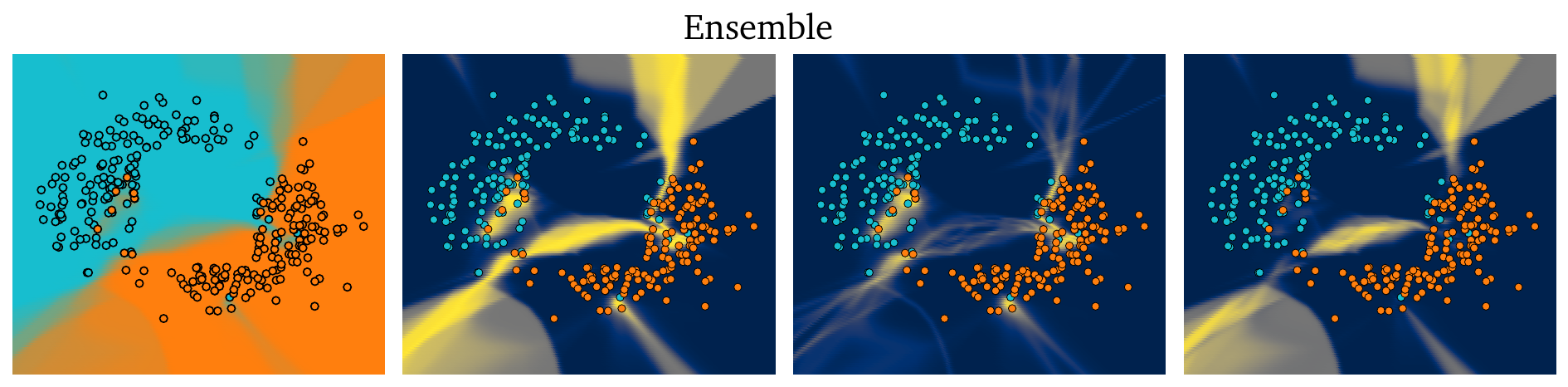}
    \includegraphics[width=0.98\linewidth]{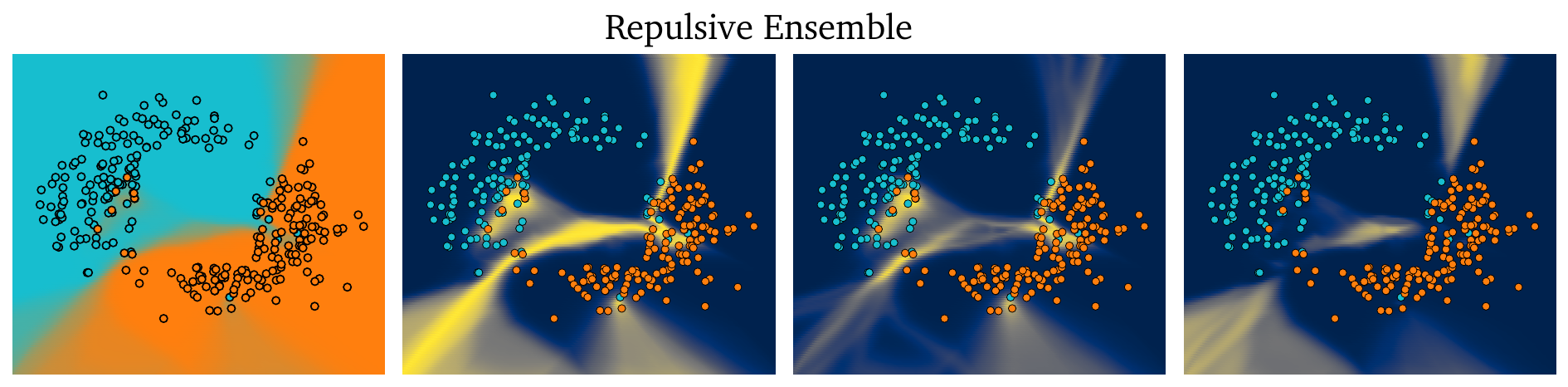}
    \includegraphics[width=0.98\linewidth]{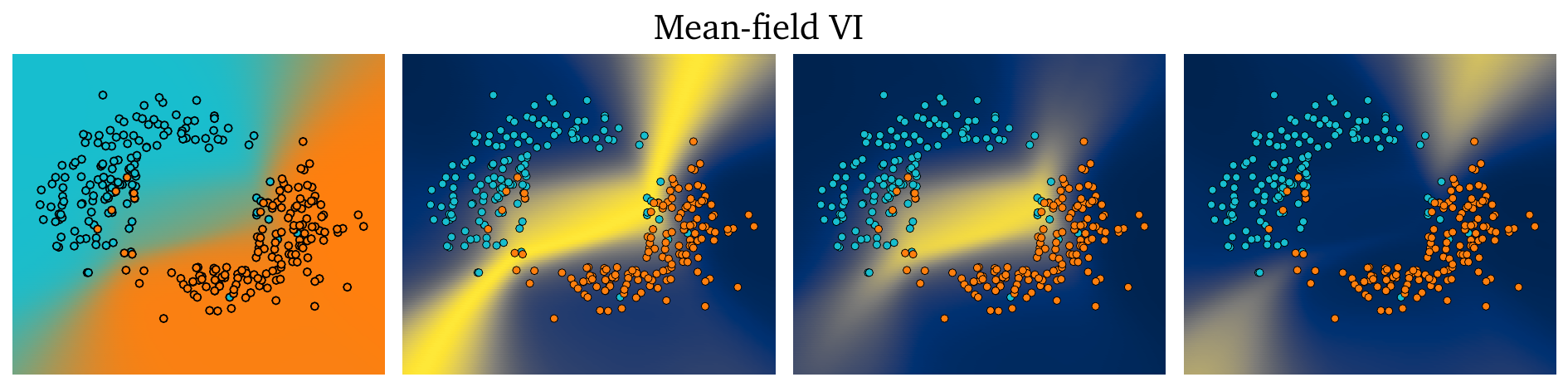}
    \includegraphics[width=0.98\linewidth]{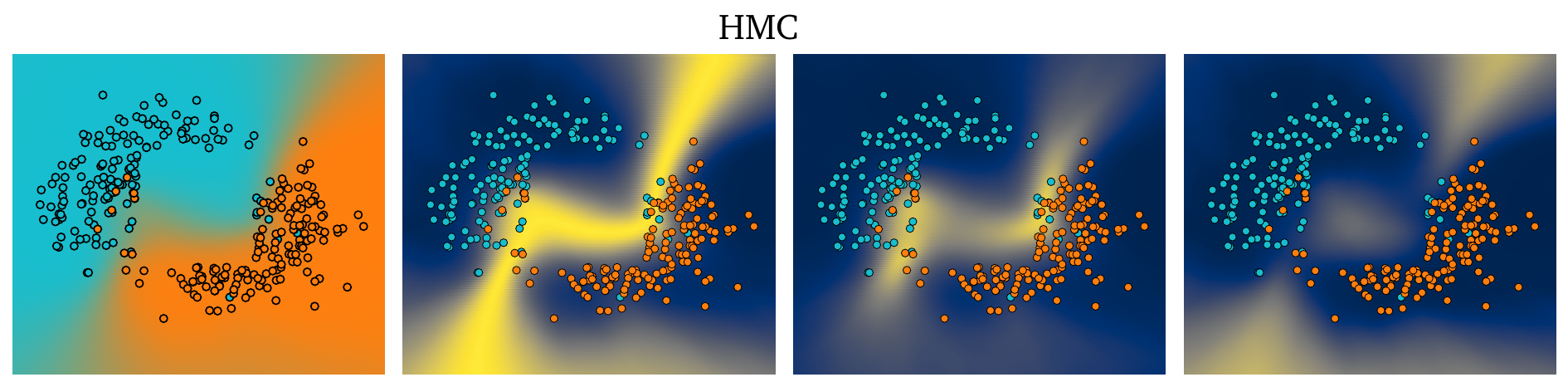}
    \caption{Binary classification example comparing five approaches. Each plot shows the probability of predicting the orange class \textit{(first column)}, the predictive entropy \textit{(second)}, the aleatoric uncertainty \textit{(third)}, and the epistemic uncertainty \textit{(fourth)}. In each plot the blue and orange dots are the observed training data. The color scales are $[0,1]$ for the first column and $[0,\log 2]$ for the others.}
    \label{fig:toy_ber}
\end{figure}

We visualize five variants of the same MLP backbone in \cref{fig:toy_ber}. The first column shows that, while the decision boundaries vary between the models in their details, they agree overall. 
Restricted to a small set of members, the predictive entropy (second column) of the two ensembles is very sharp, and its decomposition roughly separates, as expected, into an aleatoric contribution where observations from the two clusters overlap and an epistemic one where less data are observed. 
While it is reasonable to expect that each ensemble member converged to a different mode of the posterior, the BNN tends to concentrate on a single mode due to the mean-field approximation.  
However, given that it fits a multivariate normal distribution to this mode, its boundary is much smoother. Its predictive uncertainty decomposes into an aleatoric part within the overlapping data and missing square, and an epistemic contribution outside the data distribution.
The HMC samples finally provide the cleanest signal and smoothly distinguish uncertainty due to cluster overlap from uncertainty due to lack of data.

\subsubsection{Validation diagnostics}
\label{subsubsec:classification_validation}

\begin{table}[t!]
\centering
\begin{tabular}{lcc}
\toprule
Method & Brier score ($\downarrow$) & ECE ($\downarrow$)\\
\midrule
Deterministic       & $0.063 \pm 0.009$ & $0.066 \pm 0.024$\\
Ensemble            & $0.060 \pm 0.009$ & $0.052 \pm 0.019$\\
Rep. Ensemble       & $0.057 \pm 0.010$ & $0.041 \pm 0.010$\\
Mean-field VI       & $0.049 \pm 0.008$ & $0.032 \pm 0.010$\\
HMC                 & $0.047 \pm 0.007$ & $0.032 \pm 0.009$\\
\bottomrule
\end{tabular}
\caption{Brier score and Expected Calibration Error (ECE, with adaptive binning) for the binary classification example, averaged over ten random seeds $(\text{mean} \pm \text{std})$. Evaluated on held-out test sets of $1000$ points each.}
\label{tab:class_diagnostics}
\end{table}

As for the regression example, we complement the qualitative comparison in \cref{fig:toy_ber} with quantitative validation diagnostics.
Since classification outputs are probabilities rather than continuous predictions with explicit standard deviations, the natural diagnostics differ from the regression case: instead of pull distributions and coverage curves, we assess calibration via the Expected Calibration Error (ECE) (see \cref{subsec:coverage}) and the Brier score (see \cref{subsec:score}), evaluated on held-out test sets of $1000$ points drawn from the same two-moons distribution and averaged over ten random seeds.
Instead of fixed weight bins, we use quantile-based (adaptive) binning to mitigate the sensitivity to bin placement noted in \cref{subsec:score}.
Results are summarized in \cref{tab:class_diagnostics}.

Both metrics show a consistent ordering: the deterministic MLP performs worst, the two ensemble methods improve progressively with the repulsive ensemble outperforming the naive ensemble, reflecting its broader coverage of the posterior, and the two posterior inference methods achieve the best calibration.
HMC and mean-field VI are effectively indistinguishable within the reported uncertainties, despite the former being asymptotically exact and the latter restricted to a single mode with a mean-field approximation.
This suggests that for this problem, the dominant gains come from performing posterior inference at all, rather than from the fidelity of the approximation.
We caution that this conclusion is specific to the present toy setup: for posteriors with well-separated modes, strong correlation structure, or heavy tails, the gap between HMC and mean-field VI can be substantial, and the choice of approximation becomes consequential for downstream uncertainty estimates.

The ECE values show larger relative uncertainties than the Brier scores, consistent with the sensitivity of binned calibration metrics to the particular test sample.
This reinforces the point made in \cref{sec:evaluation}: while ECE provides a useful diagnostic, binning-free scoring rules such as the Brier score offer a more stable basis for comparing models, particularly when prediction distributions differ substantially across methods.

\clearpage
\section{Conclusions and outlook}
\label{sec:outlook}

In this article, we have presented a structured review of uncertainty quantification in ML for physics, with the aim of clarifying concepts, reconciling standard terminology, and highlighting statistically well-defined tools for quantification and validation.
A central organizing principle was the distinction between the \emph{source} and \emph{nature} of uncertainty, summarised by the complementary statistical--systematic and aleatoric--epistemic axes.
The former reflects the traditional classification used in experimental and theoretical physics, while the latter emphasizes reducibility and the role of limited knowledge from a statistical and ML perspective.
In addition, we have stressed that any meaningful discussion of uncertainty must specify its \emph{object}: inference uncertainty concerns parameters, functions, or model hypotheses, whereas predictive uncertainty concerns future or unobserved observables.
Making these distinctions explicit helps disentangle intrinsic randomness from limited knowledge and from modelling assumptions, and clarifies which parts of an uncertainty budget may in principle be reduced by more data, better optimisation, improved inductive biases, or refined theoretical input.

We have emphasized that the usefulness of any UQ method depends not only on its formal derivation, but also on its empirical validation.
Coverage, calibration, pull distributions, bias tests, and proper scoring rules provide a common language to assess whether uncertainty estimates are statistically consistent and neither over- nor under-confident.
At the same time, these diagnostics probe different aspects of the problem: coverage and calibration primarily test \emph{predictive} uncertainty, while bias tests probe systematic shifts in inferred quantities or functionals thereof, and proper scoring rules assess the overall fidelity of the predictive distribution.
The regression and classification examples illustrate that rather different methods -- including Gaussian processes, Bayesian neural networks, repulsive ensembles, and conformal prediction -- can behave similarly in data-rich interpolation regimes, yet lead to visibly different uncertainty estimates in extrapolation and out-of-distribution regions.
These differences are not accidental but reflect distinct inductive biases, prior assumptions, and approximation choices.

A key lesson from our worked examples is that, in extrapolation regions, calibration of uncertainty estimates becomes fundamentally dependent on modelling assumptions rather than on data. 
None of the methods we tested maintains nominal coverage uniformly outside the training support in the configurations considered, and the quoted uncertainties there reflect prior assumptions and architectural choices more than statistical constraint from observations.
While some methods -- \eg Gaussian processes with carefully chosen kernels -- can perform better than others in mild extrapolation, no method offers reliable coverage in this regime without further assumptions, and uncertainty in poorly constrained regions must be interpreted with corresponding care.

Looking ahead, many of the conceptual and technical building blocks for uncertainty quantification in physics-informed ML are already in place.
Scalable methods for high-dimensional and strongly correlated problems are increasingly available, including ensembles, Bayesian neural networks, Gaussian-process--inspired surrogates, and hybrid approaches already used in applications such as global PDF analyses and related inference problems.
Likewise, extending the present discussion from supervised prediction to fully generative models is not a conceptual break.
The same core questions reappear there in slightly different form: whether learned probability distributions are calibrated, whether derived observables achieve the expected coverage, and whether uncertainty estimates remain reliable under sampling, reweighting, and downstream inference.
These directions connect directly to companion contributions within the VERaiPHY initiative, in particular on parameter estimation and simulation-based inference~\cite{VERaiPHY_inference}, or generative modelling in general~\cite{VERaiPHY_generative}.

The most critical open issue is therefore not the availability of methods, but their robustness and interpretation under realistic and non-ideal assumptions.
Model misspecification, prior dependence, approximate inference, and dataset shift can all undermine the validity of uncertainty statements.
In particular, misspecification and distribution shift can break the assumptions under which frequentist coverage guarantees, Bayesian posterior interpretations, and empirical calibration diagnostics are usually justified.
Within the taxonomy introduced in this work, these effects should largely be understood as systematic forms of epistemic uncertainty: they do not disappear simply by increasing the sample size in the original training domain, but require revised modelling assumptions, new information, or dedicated stress tests.
This is especially relevant in precision measurements and new-physics searches, where extrapolation beyond data-constrained regions is often unavoidable and underestimated uncertainties can lead to biased inference or overconfident conclusions.
These issues are the subject of another VERaiPHY contribution focusing on robustness, stress testing, and the impact of model misspecification on downstream inference~\cite{VERaiPHY_robustness}.

Finally, uncertainty quantification is tightly connected to the broader programme of statistical inference in physics.
Reliable predictive uncertainties are a prerequisite for downstream tasks such as unfolding, simulation-based inference, PDF determination, and global fits, where they propagate into confidence intervals for physical parameters and ultimately into scientific claims.
In this sense, robust UQ is not an optional add-on, but a structural component of the full inference pipeline.
More broadly, uncertainty quantification should not be viewed as a single number or a single method, but as a property of the entire pipeline -- spanning modelling assumptions, inference procedures, approximation choices, and validation diagnostics.
Failures can occur at any of these stages, and trustworthy scientific use of ML requires that each of them be made explicit and tested.

\subsection*{Code availability and reproducibility}
\label{sec:code_availability}

All code and data needed to reproduce the results in this work are publicly available in the GitHub repository
\href{https://github.com/ramonpeter/uqveraiphy}{UQveraiphy}.
The repository contains self-contained Jupyter notebooks for each benchmark, together with shared Python utilities for data generation, model components, training, and plotting.
To make the examples broadly accessible, the repository includes implementations in multiple ML frameworks: the regression examples are based primarily on \texttt{PyTorch}~\cite{Ansel2024PyTorch2F}, with the Gaussian-process baseline implemented via \texttt{Scikit-learn}~\cite{scikit-learn}, while the classification examples are implemented in \texttt{JAX}~\cite{jax2018github} and \texttt{Equinox}~\cite{kidger2021equinox}, with HMC based on \texttt{BlackJAX}~\cite{cabezas2024blackjax}.
Running the notebooks with the default hyperparameters reproduces the figures shown in this work from scratch.

\subsection*{Acknowledgements}

We are grateful to Mikael Kuusela for many insightful and enjoyable discussions on the statistical aspects of this work, and for his thoughtful comments on an early version of the manuscript.
We also thank Rikab Gambhir and Samuel Klein their careful review of the draft and for their valuable feedback.
We are particularly thankful to Gaia Grosso for coordinating the VERaiPHY initiative, fostering communication across the different teams, and leading many fruitful discussions at the workshop.
We warmly thank Tilman Plehn, Lydia Brenner, and Louis Lyons for initiating this effort in the first place.
More broadly, we thank Lydia Brenner and Louis Lyons for driving the broader PHYSTAT effort, and for helping anchor the VERaiPHY initiative within it. MU is supported by the European Research Council under the European Union’s Horizon 2020 research and innovation Programme (PBSP, Grant agreement
n.950246) and partially supported by the STFC consolidated grant ST/X000664/1.

\clearpage

\bibliography{refs}

\providecommand{\href}[2]{#2}\begingroup\raggedright\begin{thebibliography}{100}

\bibitem{stat_phys_2006}
F.~James, \href{http://dx.doi.org/10.1142/6096}{{\em Statistical Methods in Experimental Physics}}.
\newblock World Scientific, 2nd~ed., 2006.

\bibitem{Trotta:2008qt}
R.~Trotta, {\it {Bayes in the sky: Bayesian inference and model selection in cosmology}},  \href{http://dx.doi.org/10.1080/00107510802066753}{Contemp. Phys. {\bfseries 49} (2008)  71}, \href{http://arxiv.org/abs/0803.4089}{{arXiv:0803.4089 [astro-ph]}}.

\bibitem{hogg2010astro}
D.~W. Hogg, J.~Bovy, and D.~Lang, {\it Data analysis recipes: Fitting a model to data},  \href{http://arxiv.org/abs/1008.4686}{{arXiv:1008.4686 [astro-ph.IM]}}.

\bibitem{Cowan:2010js}
G.~Cowan, K.~Cranmer, E.~Gross, and O.~Vitells, {\it {Asymptotic formulae for likelihood-based tests of new physics}},  \href{http://dx.doi.org/10.1140/epjc/s10052-011-1554-0}{Eur. Phys. J. C {\bfseries 71} (2011)  1554}, \href{http://arxiv.org/abs/1007.1727}{{arXiv:1007.1727 [physics.data-an]}}. [Erratum: Eur.Phys.J.C 73, 2501 (2013)].

\bibitem{Trotta:2017wnx}
R.~Trotta, {\it {Bayesian Methods in Cosmology}},  \href{http://arxiv.org/abs/1701.01467}{{arXiv:1701.01467 [astro-ph.CO]}}.

\bibitem{Kendall2017_uq}
A.~Kendall and Y.~Gal, {\it What uncertainties do we need in bayesian deep learning for computer vision?},  in {\em Advances in Neural Information Processing Systems}.
\newblock Curran Associates, Inc., 2017.
\newblock \href{http://arxiv.org/abs/1703.04977}{{arXiv:1703.04977 [cs.CV]}}.

\bibitem{MLgenUQ1}
J.~Gawlikowski, C.~Tassi, and M.~e.~a. Ali, {\it A survey of uncertainty in deep neural networks},  \href{http://dx.doi.org/s10462-023-10562-9}{Artif Intell Rev 56 {\bfseries (Suppl 1)} (2023)  1513–1589}, \href{http://arxiv.org/abs/2107.03342}{{arXiv:2107.03342 [cs.LG]}}.

\bibitem{MLgenUQ2}
F.~Fakour, A.~Mosleh, and R.~Ramezani, {\it A structured review of literature on uncertainty in machine learning \& deep learning},  \href{http://arxiv.org/abs/2406.00332}{{arXiv:2406.00332 [cs.LG]}}.

\bibitem{MLgenUQ3}
N.~Stahl, G.~Falkman, A.~Karlsson, and G.~Mathiason, \href{http://dx.doi.org/10.1007/978-3-030-50146-4_41}{{\it Evaluation of uncertainty quantification in deep learning.}, } in {\em Information Processing and Management of Uncertainty in Knowledge-Based Systems}.
\newblock Springer International Publishing, 2020.

\bibitem{Abdar_2021}
M.~Abdar, F.~Pourpanah, S.~Hussain, D.~Rezazadegan, L.~Liu, M.~Ghavamzadeh, P.~Fieguth, X.~Cao, A.~Khosravi, U.~R. Acharya, V.~Makarenkov, and S.~Nahavandi, {\it A review of uncertainty quantification in deep learning: Techniques, applications and challenges},  \href{http://dx.doi.org/10.1016/j.inffus.2021.05.008}{Information Fusion {\bfseries 76} (Dec., 2021)   243–297}.

\bibitem{PHYgenUQ1}
T.~Siddique, M.~S. Mahmud, A.~M. Keesee, C.~M. Ngwira, and H.~Connor, {\it A survey of uncertainty quantification in machine learning for space weather prediction},  \href{http://dx.doi.org/10.3390/geosciences12010027}{Geosciences {\bfseries 12} (2022) 1, }.

\bibitem{Kompa2021_uncertainty_medical}
B.~Kompa, J.~Snoek, and A.~L. Beam, {\it Second opinion needed: communicating uncertainty in medical machine learning},  \href{http://dx.doi.org/10.1038/s41746-020-00367-3}{npj Digital Medicine {\bfseries 4} (2021) 1, 4}. \href{https://doi.org/10.1038/s41746-020-00367-3}{{\tt https://doi.org/10.1038/s41746-020-00367-3}}.

\bibitem{Loftus2022_uncertainty_healthcare}
T.~J. Loftus, B.~Shickel, M.~M. Ruppert, J.~A. Balch, T.~Ozrazgat-Baslanti, P.~J. Tighe, P.~A. Efron, W.~R. Hogan, P.~Rashidi, G.~R. Upchurch, Jr., and A.~Bihorac, {\it Uncertainty-aware deep learning in healthcare: A scoping review},  \href{http://dx.doi.org/10.1371/journal.pdig.0000085}{PLOS Digital Health {\bfseries 1} (08, 2022)   1}. \href{https://doi.org/10.1371/journal.pdig.0000085}{{\tt https://doi.org/10.1371/journal.pdig.0000085}}.

\bibitem{Araujo2024_autonomous_driving_uncertainty}
B.~Araújo, J.~F. Teixeira, J.~Fonseca, R.~Cerqueira, and S.~C. Beco, {\it The road to safety: A review of uncertainty and applications to autonomous driving perception},  \href{http://dx.doi.org/10.3390/e26080634}{Entropy {\bfseries 26} (2024) 8, }. \href{https://www.mdpi.com/1099-4300/26/8/634}{{\tt https://www.mdpi.com/1099-4300/26/8/634}}.

\bibitem{pmlr-v70-guo17a}
C.~Guo, G.~Pleiss, Y.~Sun, and K.~Q. Weinberger, {\it On calibration of modern neural networks},  in {\em Proceedings of the 34th International Conference on Machine Learning}, D.~Precup and Y.~W. Teh, eds.
\newblock PMLR, 06--11 Aug, 2017.
\newblock \href{http://arxiv.org/abs/1706.04599}{{arXiv:1706.04599 [cs.LG]}}.

\bibitem{Forte:2002fg}
S.~Forte, L.~Garrido, J.~I. Latorre, and A.~Piccione, {\it {Neural network parametrization of deep inelastic structure functions}},  \href{http://dx.doi.org/10.1088/1126-6708/2002/05/062}{JHEP {\bfseries 05} (2002)  062}, \href{http://arxiv.org/abs/hep-ph/0204232}{{arXiv:hep-ph/0204232}}.

\bibitem{Ball:2008by}
NNPDF, R.~D. Ball, L.~Del~Debbio, S.~Forte, A.~Guffanti, J.~I. Latorre, A.~Piccione, J.~Rojo, and M.~Ubiali, {\it {A Determination of parton distributions with faithful uncertainty estimation}},  \href{http://dx.doi.org/10.1016/j.nuclphysb.2008.09.037}{Nucl. Phys. B {\bfseries 809} (2009)  1}, \href{http://arxiv.org/abs/0808.1231}{{arXiv:0808.1231 [hep-ph]}}. [Erratum: Nucl.Phys.B 816, 293 (2009)].

\bibitem{Englert:2018cfo}
C.~Englert, P.~Galler, P.~Harris, and M.~Spannowsky, {\it {Machine Learning Uncertainties with Adversarial Neural Networks}},  \href{http://dx.doi.org/10.1140/epjc/s10052-018-6511-8}{Eur. Phys. J. C {\bfseries 79} (2019) 1, 4}, \href{http://arxiv.org/abs/1807.08763}{{arXiv:1807.08763 [hep-ph]}}.

\bibitem{Bollweg:2019skg}
S.~Bollweg, M.~Hau{\ss}mann, G.~Kasieczka, M.~Luchmann, T.~Plehn, and J.~Thompson, {\it {Deep-Learning Jets with Uncertainties and More}},  \href{http://dx.doi.org/10.21468/SciPostPhys.8.1.006}{SciPost Phys. {\bfseries 8} (2020) 1, 006}, \href{http://arxiv.org/abs/1904.10004}{{arXiv:1904.10004 [hep-ph]}}.

\bibitem{Carrazza:2019mzf}
S.~Carrazza and J.~Cruz-Martinez, {\it {Towards a new generation of parton densities with deep learning models}},  \href{http://dx.doi.org/10.1140/epjc/s10052-019-7197-2}{Eur. Phys. J. C {\bfseries 79} (2019) 8, 676}, \href{http://arxiv.org/abs/1907.05075}{{arXiv:1907.05075 [hep-ph]}}.

\bibitem{Nachman:2019dol}
B.~Nachman, {\it {A guide for deploying Deep Learning in LHC searches: How to achieve optimality and account for uncertainty}},  \href{http://dx.doi.org/10.21468/SciPostPhys.8.6.090}{SciPost Phys. {\bfseries 8} (2020)  090}, \href{http://arxiv.org/abs/1909.03081}{{arXiv:1909.03081 [hep-ph]}}.

\bibitem{Kasieczka:2020vlh}
G.~Kasieczka, M.~Luchmann, F.~Otterpohl, and T.~Plehn, {\it {Per-Object Systematics using Deep-Learned Calibration}},  \href{http://dx.doi.org/10.21468/SciPostPhys.9.6.089}{SciPost Phys. {\bfseries 9} (2020)  089}, \href{http://arxiv.org/abs/2003.11099}{{arXiv:2003.11099 [hep-ph]}}.

\bibitem{Araz:2021wqm}
J.~Y. Araz and M.~Spannowsky, {\it {Combine and Conquer: Event Reconstruction with Bayesian Ensemble Neural Networks}},  \href{http://dx.doi.org/10.1007/JHEP04(2021)296}{JHEP {\bfseries 04} (2021)  296}, \href{http://arxiv.org/abs/2102.01078}{{arXiv:2102.01078 [hep-ph]}}.

\bibitem{Bellagente:2021yyh}
M.~Bellagente, M.~Haussmann, M.~Luchmann, and T.~Plehn, {\it {Understanding Event-Generation Networks via Uncertainties}},  \href{http://dx.doi.org/10.21468/SciPostPhys.13.1.003}{SciPost Phys. {\bfseries 13} (4, 2022)   003}, \href{http://arxiv.org/abs/2104.04543}{{arXiv:2104.04543 [hep-ph]}}.

\bibitem{Ghosh:2021hrh}
A.~Ghosh and B.~Nachman, {\it {A cautionary tale of decorrelating theory uncertainties}},  \href{http://dx.doi.org/10.1140/epjc/s10052-022-10012-w}{Eur. Phys. J. C {\bfseries 82} (2022) 1, 46}, \href{http://arxiv.org/abs/2109.08159}{{arXiv:2109.08159 [hep-ph]}}.

\bibitem{Ghosh:2021roe}
A.~Ghosh, B.~Nachman, and D.~Whiteson, {\it {Uncertainty-aware machine learning for high energy physics}},  \href{http://dx.doi.org/10.1103/PhysRevD.104.056026}{Phys. Rev. D {\bfseries 104} (2021) 5, 056026}, \href{http://arxiv.org/abs/2105.08742}{{arXiv:2105.08742 [physics.data-an]}}.

\bibitem{NNPDF:2021njg}
NNPDF, R.~D. Ball {\em et al.}, {\it {The path to proton structure at 1{\%} accuracy}},  \href{http://dx.doi.org/10.1140/epjc/s10052-022-10328-7}{Eur. Phys. J. C {\bfseries 82} (2022) 5, 428}, \href{http://arxiv.org/abs/2109.02653}{{arXiv:2109.02653 [hep-ph]}}.

\bibitem{NNPDF:2021uiq}
NNPDF, R.~D. Ball {\em et al.}, {\it {An open-source machine learning framework for global analyses of parton distributions}},  \href{http://dx.doi.org/10.1140/epjc/s10052-021-09747-9}{Eur. Phys. J. C {\bfseries 81} (2021) 10, 958}, \href{http://arxiv.org/abs/2109.02671}{{arXiv:2109.02671 [hep-ph]}}.

\bibitem{Badger:2022hwf}
S.~Badger, A.~Butter, M.~Luchmann, S.~Pitz, and T.~Plehn, {\it {Loop amplitudes from precision networks}},  \href{http://dx.doi.org/10.21468/SciPostPhysCore.6.2.034}{SciPost Phys. Core {\bfseries 6} (2023)  034}, \href{http://arxiv.org/abs/2206.14831}{{arXiv:2206.14831 [hep-ph]}}.

\bibitem{Chen:2022pzc}
T.~Y. Chen, B.~Dey, A.~Ghosh, M.~Kagan, B.~Nord, and N.~Ramachandra, \href{http://dx.doi.org/10.2172/1886020}{{\it {Interpretable Uncertainty Quantification in AI for HEP}}, } in {\em {Snowmass 2021}}.
\newblock 8, 2022.
\newblock \href{http://arxiv.org/abs/2208.03284}{{arXiv:2208.03284 [hep-ex]}}.

\bibitem{Butter:2022vkj}
A.~Butter, T.~Heimel, T.~Martini, S.~Peitzsch, and T.~Plehn, {\it {Two invertible networks for the matrix element method}},  \href{http://dx.doi.org/10.21468/SciPostPhys.15.3.094}{SciPost Phys. {\bfseries 15} (2023) 3, 094}, \href{http://arxiv.org/abs/2210.00019}{{arXiv:2210.00019 [hep-ph]}}.

\bibitem{Fanelli:2023lmp}
C.~Fanelli and J.~Giroux, {\it {ELUQuant: event-level uncertainty quantification in deep inelastic scattering}},  \href{http://dx.doi.org/10.1088/2632-2153/ad2098}{Mach. Learn. Sci. Tech. {\bfseries 5} (2024) 1, 015017}, \href{http://arxiv.org/abs/2310.02913}{{arXiv:2310.02913 [cs.LG]}}.

\bibitem{Heimel:2023mvw}
T.~Heimel, N.~Huetsch, R.~Winterhalder, T.~Plehn, and A.~Butter, {\it {Precision-machine learning for the matrix element method}},  \href{http://dx.doi.org/10.21468/SciPostPhys.17.5.129}{SciPost Phys. {\bfseries 17} (2024) 5, 129}, \href{http://arxiv.org/abs/2310.07752}{{arXiv:2310.07752 [hep-ph]}}.

\bibitem{ATLAS:2024rpl}
{ATLAS Collaboration}, {\it {Precision calibration of calorimeter signals in the ATLAS experiment using an uncertainty-aware neural network}},  \href{http://dx.doi.org/10.21468/SciPostPhys.19.6.155}{SciPost Phys. {\bfseries 19} (2025) 6, 155}, \href{http://arxiv.org/abs/2412.04370}{{arXiv:2412.04370 [hep-ex]}}.

\bibitem{Kriesten:2024ist}
B.~Kriesten and T.~J. Hobbs, {\it {Anomalous electroweak physics unraveled via evidential deep learning}},  \href{http://dx.doi.org/10.1140/epjc/s10052-025-14501-6}{Eur. Phys. J. C {\bfseries 85} (2025) 8, 883}, \href{http://arxiv.org/abs/2412.16286}{{arXiv:2412.16286 [hep-ph]}}.

\bibitem{Benato:2024lnj}
L.~Benato {\em et al.}, {\it {FAIR Universe HiggsML Uncertainty Dataset and Competition}},  \href{http://arxiv.org/abs/2410.02867}{{arXiv:2410.02867 [hep-ph]}}.

\bibitem{Huetsch:2024quz}
N.~Huetsch {\em et al.}, {\it {The landscape of unfolding with machine learning}},  \href{http://dx.doi.org/10.21468/SciPostPhys.18.2.070}{SciPost Phys. {\bfseries 18} (2025) 2, 070}, \href{http://arxiv.org/abs/2404.18807}{{arXiv:2404.18807 [hep-ph]}}.

\bibitem{Bieringer:2024nbc}
S.~Bieringer, S.~Diefenbacher, G.~Kasieczka, and M.~Trabs, {\it {Calibrating Bayesian generative machine learning for Bayesiamplification}},  \href{http://dx.doi.org/10.1088/2632-2153/ad9136}{Mach. Learn. Sci. Tech. {\bfseries 5} (2024) 4, 045044}, \href{http://arxiv.org/abs/2408.00838}{{arXiv:2408.00838 [cs.LG]}}.

\bibitem{Bahl:2024gyt}
H.~Bahl, N.~Elmer, L.~Favaro, M.~Haussmann, T.~Plehn, and R.~Winterhalder, {\it {Accurate surrogate amplitudes with calibrated uncertainties}},  \href{http://dx.doi.org/10.21468/SciPostPhysCore.8.4.073}{SciPost Phys. Core {\bfseries 8} (2025)  073}, \href{http://arxiv.org/abs/2412.12069}{{arXiv:2412.12069 [hep-ph]}}.

\bibitem{Benevedes:2025nzr}
S.~Benevedes and J.~Thaler, {\it {Frequentist uncertainties on neural density ratios with wi{\,}fi ensembles}},  \href{http://dx.doi.org/10.1103/w28w-x5wh}{Phys. Rev. D {\bfseries 112} (2025) 5, 056024}, \href{http://arxiv.org/abs/2506.00113}{{arXiv:2506.00113 [hep-ph]}}.

\bibitem{Khot:2025kqg}
A.~Khot, X.~Wang, A.~Roy, V.~Kindratenko, and M.~S. Neubauer, {\it {Evidential deep learning for uncertainty quantification and out-of-distribution detection in jet identification using deep neural networks}},  \href{http://dx.doi.org/10.1088/2632-2153/ade51b}{Mach. Learn. Sci. Tech. {\bfseries 6} (2025) 3, 035003}, \href{http://arxiv.org/abs/2501.05656}{{arXiv:2501.05656 [hep-ex]}}.

\bibitem{Bahl:2025xvx}
H.~Bahl, N.~Elmer, T.~Plehn, and R.~Winterhalder, {\it {Amplitude Uncertainties Everywhere All at Once}},  \href{http://dx.doi.org/10.21468/SciPostPhys.20.3.083}{SciPost Phys. {\bfseries 20} (2026)  083}, \href{http://arxiv.org/abs/2509.00155}{{arXiv:2509.00155 [hep-ph]}}.

\bibitem{Bendavid:2025urn}
J.~Bendavid, D.~Conde, M.~Morales-Alvarado, V.~Sanz, and M.~Ubiali, {\it {Angular coefficients from interpretable machine learning with symbolic regression}},  \href{http://dx.doi.org/10.1007/JHEP02(2026)081}{JHEP {\bfseries 02} (2026)  081}, \href{http://arxiv.org/abs/2508.00989}{{arXiv:2508.00989 [hep-ph]}}.

\bibitem{Araz:2025vuw}
J.~Y. Araz and M.~Spannowsky, {\it {Another Fit Bites the Dust: Conformal Prediction as a Calibration Standard for Machine Learning in High-Energy Physics}},  \href{http://arxiv.org/abs/2512.17048}{{arXiv:2512.17048 [hep-ph]}}.

\bibitem{Costantini:2025wxp}
M.~N. Costantini, L.~Mantani, J.~M. Moore, and M.~Ubiali, {\it {A linear PDF model for Bayesian inference}},  \href{http://dx.doi.org/10.1007/JHEP04(2026)068}{JHEP {\bfseries 04} (2026)  068}, \href{http://arxiv.org/abs/2507.16913}{{arXiv:2507.16913 [hep-ph]}}.

\bibitem{Costantini:2025agd}
M.~N. Costantini, L.~Mantani, J.~M. Moore, V.~S. S{\'a}nchez, and M.~Ubiali, {\it {Colibri: A new tool for fast-flying PDF fits}},  \href{http://dx.doi.org/10.1140/epjc/s10052-025-15185-8}{Eur. Phys. J. C {\bfseries 86} (2026) 1, 22}, \href{http://arxiv.org/abs/2510.03391}{{arXiv:2510.03391 [hep-ph]}}.

\bibitem{Beccatini:2025tpk}
L.~Beccatini, F.~Maltoni, O.~Mattelaer, and R.~Winterhalder, {\it {Amplitude Surrogates for Multi-Jet Processes}},  \href{http://arxiv.org/abs/2512.11036}{{arXiv:2512.11036 [hep-ph]}}.

\bibitem{Benato:2025rgo}
L.~Benato, C.~Giordano, C.~Krause, A.~Li, R.~Sch{\"o}fbeck, D.~Schwarz, M.~Shooshtari, and D.~Wang, {\it {Unbinned inclusive cross-section measurements with machine-learned systematic uncertainties}},  \href{http://dx.doi.org/10.1103/zwzt-1rrw}{Phys. Rev. D {\bfseries 112} (2025) 5, 052006}, \href{http://arxiv.org/abs/2505.05544}{{arXiv:2505.05544 [hep-ph]}}.

\bibitem{Barontini:2025lnl}
A.~Barontini, M.~N. Costantini, G.~De~Crescenzo, S.~Forte, and M.~Ubiali, {\it {Evaluating the faithfulness of PDF uncertainties in the presence of inconsistent data}},  \href{http://arxiv.org/abs/2503.17447}{{arXiv:2503.17447 [hep-ph]}}.

\bibitem{Bahl:2026qaf}
H.~Bahl, J.~Braun, G.~Heinrich, T.~Plehn, and R.~Revelli, {\it {How to Trust Learned Loop Amplitudes}},  \href{http://arxiv.org/abs/2601.00950}{{arXiv:2601.00950 [hep-ph]}}.

\bibitem{Padilla:2019mgi}
L.~E. Padilla, L.~O. Tellez, L.~A. Escamilla, and J.~A. Vazquez, {\it {Cosmological Parameter Inference with Bayesian Statistics}},  \href{http://dx.doi.org/10.3390/universe7070213}{Universe {\bfseries 7} (2021) 7, 213}, \href{http://arxiv.org/abs/1903.11127}{{arXiv:1903.11127 [astro-ph.CO]}}.

\bibitem{Rover:2024pvr}
L.~R{\"o}ver, B.~M. Sch{\"a}fer, and T.~Plehn, {\it {PINNferring the Hubble Function with Uncertainties}},  \href{http://arxiv.org/abs/2403.13899}{{arXiv:2403.13899 [astro-ph.CO]}}.

\bibitem{Herold:2024enb}
L.~Herold, E.~G.~M. Ferreira, and L.~Heinrich, {\it {Profile likelihoods in cosmology: When, why, and how illustrated with {\ensuremath{\Lambda}}CDM, massive neutrinos, and dark energy}},  \href{http://dx.doi.org/10.1103/PhysRevD.111.083504}{Phys. Rev. D {\bfseries 111} (2025) 8, 083504}, \href{http://arxiv.org/abs/2408.07700}{{arXiv:2408.07700 [astro-ph.CO]}}.

\bibitem{Drabicki:2026rgn}
K.~Drabicki, S.~J. Nakoneczny, and M.~Bilicki, {\it {Modeling Quasar Photo-$z$ Distribution and Uncertainty. A Study Based on the Kilo-Degree Survey}},  \href{http://arxiv.org/abs/2603.19882}{{arXiv:2603.19882 [astro-ph.CO]}}.

\bibitem{Soriano:2026mwu}
J.~Soriano, T.~Do, S.~Saikrishnan, V.~Seenivasan, B.~Boscoe, J.~Singal, and E.~Jones, {\it {Improving Generalization and Uncertainty Quantification of Photometric Redshift Models}},  \href{http://dx.doi.org/10.3847/1538-3881/ae2ffe}{Astron. J. {\bfseries 171} (2026) 2, 114}, \href{http://arxiv.org/abs/2601.17222}{{arXiv:2601.17222 [astro-ph.IM]}}.

\bibitem{Leterme:2026cqj}
H.~Leterme, A.~Tersenov, J.~Fadili, and J.-L. Starck, {\it {A plug-and-play approach with fast uncertainty quantification for weak lensing mass mapping}},  \href{http://arxiv.org/abs/2603.22006}{{arXiv:2603.22006 [astro-ph.CO]}}.

\bibitem{Dai:2026jme}
B.~Dai {\em et al.}, {\it {FAIR Universe Weak Lensing ML Uncertainty Challenge: Handling Uncertainties and Distribution Shifts for Precision Cosmology}},  \href{http://arxiv.org/abs/2604.14451}{{arXiv:2604.14451 [astro-ph.CO]}}.

\bibitem{Butter:2021mwl}
A.~Butter, T.~Finke, F.~Keil, M.~Kr{\"a}mer, and S.~Manconi, {\it {Classification of Fermi-LAT blazars with Bayesian neural networks}},  \href{http://dx.doi.org/10.1088/1475-7516/2022/04/023}{JCAP {\bfseries 04} (2022) 04, 023}, \href{http://arxiv.org/abs/2112.01403}{{arXiv:2112.01403 [astro-ph.HE]}}.

\bibitem{Malz:2024zjd}
A.-K. Malz, G.~Ashton, and N.~Colombo, {\it {Classification uncertainty for transient gravitational-wave noise artifacts with optimized conformal prediction}},  \href{http://dx.doi.org/10.1103/PhysRevD.111.084078}{Phys. Rev. D {\bfseries 111} (2025) 8, 084078}, \href{http://arxiv.org/abs/2412.11801}{{arXiv:2412.11801 [gr-qc]}}.

\bibitem{Ashton:2024wae}
G.~Ashton, N.~Colombo, I.~Harry, and S.~Sachdev, {\it {Calibrating gravitational-wave search algorithms with conformal prediction}},  \href{http://dx.doi.org/10.1103/PhysRevD.109.123027}{Phys. Rev. D {\bfseries 109} (2024) 12, 123027}, \href{http://arxiv.org/abs/2402.19313}{{arXiv:2402.19313 [gr-qc]}}.

\bibitem{Sun:2024ywb}
T.-Y. Sun, Y.~Shao, Y.~Li, Y.~Xu, H.~Wang, and X.~Zhang, {\it {Deep learning-driven likelihood-free parameter inference for 21-cm forest observations}},  \href{http://dx.doi.org/10.1038/s42005-025-02139-5}{Commun. Phys. {\bfseries 8} (2025) 1, 220}, \href{http://arxiv.org/abs/2407.14298}{{arXiv:2407.14298 [astro-ph.CO]}}.

\bibitem{Perez-Romero:2025gwg}
J.~P{\'e}rez-Romero {\em et al.}, {\it {Towards a foundation model for astrophysical source detection: An End-to-End Gamma-Ray Data Analysis Pipeline Using Deep Learning}},  in {\em {2nd European AI for Fundamental Physics Conference}}.
\newblock 9, 2025.
\newblock \href{http://arxiv.org/abs/2509.25128}{{arXiv:2509.25128 [astro-ph.IM]}}.

\bibitem{Liu:2026nhu}
S.-T. Liu, T.-Y. Sun, Y.-X. Wang, Y.-X. Zhang, S.-J. Jin, J.-F. Zhang, and X.~Zhang, {\it {Assessing the robustness of amortized simulation-based inference to transient noise in gravitational-wave ringdowns}},  \href{http://arxiv.org/abs/2603.12032}{{arXiv:2603.12032 [gr-qc]}}.

\bibitem{walls2026montecarloconformalprediction}
A.~Walls, J.~Barry, D.~Mohan, and A.~M.~M. Scaife, {\it Monte carlo conformal prediction for quantifying uncertainty in radio galaxy classification under ambiguous ground truth},  \href{http://arxiv.org/abs/2603.20000}{{arXiv:2603.20000 [astro-ph.IM]}}.

\bibitem{Mendes:2026fsj}
M.~M.~S. Mendes, R.~D. Pereira, M.~D. d.~R. Louren, and C.~H. Lenzi, {\it {Certified Uncertainty for Surrogate Models of Neutron Star Equations of State via Mondrian Conformal Prediction}},  \href{http://arxiv.org/abs/2602.19363}{{arXiv:2602.19363 [astro-ph.HE]}}.

\bibitem{hong2026uncertaintyawaresolarflareregression}
J.~Hong, C.~Pandey, and B.~Aydin, {\it Uncertainty-aware solar flare regression},  \href{http://arxiv.org/abs/2603.06712}{{arXiv:2603.06712 [astro-ph.SR]}}.

\bibitem{VERaiPHY_inference}
M.~Dax, T.~Heimel, and G.~Louppe, {\it {Simulation-based Inference with Machine Learning}},  to appear (2026)  , \href{http://arxiv.org/abs/2026.xxxxx}{{arXiv:2026.xxxxx}}.

\bibitem{vovk2005algorithmic}
V.~Vovk, A.~Gammerman, and G.~Shafer, {\em Algorithmic learning in a random world}.
\newblock Springer, 2005.

\bibitem{shafer2007tutorialconformalprediction}
G.~Shafer and V.~Vovk, {\it A tutorial on conformal prediction},  \href{http://arxiv.org/abs/0706.3188}{{arXiv:0706.3188 [cs.LG]}}.

\bibitem{lakshminarayanan2017}
B.~Lakshminarayanan, A.~Pritzel, and C.~Blundell, {\it Simple and scalable predictive uncertainty estimation using deep ensembles},  \href{http://arxiv.org/abs/1612.01474}{{arXiv:1612.01474 [stat.ML]}}.

\bibitem{Gneiting01032007}
T.~Gneiting and A.~E. Raftery, {\it Strictly proper scoring rules, prediction, and estimation},  \href{http://dx.doi.org/10.1198/016214506000001437}{Journal of the American Statistical Association {\bfseries 102} (2007) 477, 359}.

\bibitem{VERaiPHY_intro}
G.~Grosso, R.~Winterhalder, L.~Brenner, L.~Lyons, and T.~Plehn, {\it {VERaiPHY -- Validation \& Evaluation for Robust AI in PHYsics}},  to appear (2026)  , \href{http://arxiv.org/abs/2026.xxxxx}{{arXiv:2026.xxxxx}}.

\bibitem{VERaiPHY_hypo}
O.~Amram, M.~Letizia, and M.~Kuusela, {\it {Model-Agnostic Signal Discovery with Machine Learning: Bridging the Gap Between Theory and Practice}},  to appear (2026)  , \href{http://arxiv.org/abs/2026.xxxxx}{{arXiv:2026.xxxxx}}.

\bibitem{DerKiureghian2009}
A.~Der~Kiureghian and O.~Ditlevsen, {\it Aleatory or epistemic? {D}oes it matter?},  \href{http://dx.doi.org/10.1016/j.strusafe.2008.06.020}{Structural Safety {\bfseries 31} (2009) 2, 105}.

\bibitem{Hullermeier2021}
E.~H{\"u}llermeier and W.~Waegeman, {\it Aleatoric and epistemic uncertainty in machine learning: an introduction to concepts and methods},  \href{http://dx.doi.org/10.1007/s10994-021-05946-3}{Machine Learning {\bfseries 110} (2021)  457}, \href{http://arxiv.org/abs/1910.09457}{{arXiv:1910.09457 [cs.LG]}}.

\bibitem{Gruber2023}
C.~Gruber, P.~O. Schenk, M.~Schierholz, F.~Kreuter, and G.~Kauermann, {\it Sources of uncertainty in supervised machine learning -- a statisticians' view},  \href{http://arxiv.org/abs/2305.16703}{{arXiv:2305.16703 [stat.ML]}}.

\bibitem{smith2025rethinking}
F.~B. Smith, J.~Kossen, E.~Trollope, M.~van~der Wilk, A.~Foster, and T.~Rainforth, {\it Rethinking aleatoric and epistemic uncertainty},  \href{http://arxiv.org/abs/2412.20892}{{arXiv:2412.20892 [cs.LG]}}.

\bibitem{epistemicuncertaintyestimation}
S.~Jiménez, M.~Jürgens, and W.~Waegeman, {\it Position: Epistemic uncertainty estimation methods are fundamentally incomplete},  \href{http://arxiv.org/abs/2505.23506}{{arXiv:2505.23506 [cs.LG]}}.

\bibitem{casella2024statistical}
G.~Casella and R.~Berger, {\em Statistical inference}.
\newblock Chapman and Hall/CRC, 2024.

\bibitem{Wasserman2004}
L.~Wasserman, \href{http://dx.doi.org/10.1007/978-0-387-21736-9}{{\em All of Statistics}}.
\newblock Springer New York, 7, 2004.

\bibitem{Angelopoulos_2021}
A.~N. Angelopoulos and S.~Bates, {\it Conformal prediction: A gentle introduction},  \href{http://dx.doi.org/10.1561/2200000101}{Foundations and Trends® in Machine Learning {\bfseries 16} (2023) 4, }, \href{http://arxiv.org/abs/2107.07511}{{arXiv:2107.07511 [cs.LG]}}.

\bibitem{valiant1984theory}
L.~G. Valiant, {\it A theory of the learnable},  Communications of the ACM {\bfseries 27} (1984) 11, 1134.

\bibitem{mcallester1998some}
D.~A. McAllester, {\it Some pac-bayesian theorems},  in {\em Proceedings of the eleventh annual conference on Computational learning theory}.
\newblock 1998.

\bibitem{kldivergence}
S.~Kullback and R.~A. Leibler, {\it On information and sufficiency},  The Annals of Mathematical Statistics {\bfseries 22} (1951) 1, 79. \href{http://www.jstor.org/stable/2236703}{{\tt http://www.jstor.org/stable/2236703}}.

\bibitem{germain2016pac}
P.~Germain, F.~Bach, A.~Lacoste, and S.~Lacoste-Julien, {\it {PAC}-{B}ayesian theory meets {B}ayesian inference},  in {\em Advances in Neural Information Processing Systems}.
\newblock 2016.

\bibitem{alquier2024user}
P.~Alquier, {\it User-friendly introduction to pac-bayes bounds},  \href{http://dx.doi.org/10.1561/2200000100}{Foundations and Trends® in Machine Learning {\bfseries 17} (2024) 2, 174–303}.

\bibitem{gelman1995bayesian}
A.~Gelman, J.~B. Carlin, H.~S. Stern, and D.~B. Rubin, {\em Bayesian data analysis}.
\newblock Chapman and Hall/CRC, 1995.

\bibitem{mcelreath2018statistical}
R.~McElreath, {\em Statistical rethinking: A Bayesian course with examples in R and Stan}.
\newblock Chapman and Hall/CRC, 2018.

\bibitem{berger2024objective}
J.~O. Berger, J.-m. Bernardo, and D.~Sun, {\em Objective Bayesian Inference}.
\newblock World Scientific, 2024.

\bibitem{jaynes2003probability}
E.~T. Jaynes, {\em Probability theory: The logic of science}.
\newblock Cambridge university press, 2003.

\bibitem{efron2012large}
B.~Efron, {\em Large-scale inference: empirical Bayes methods for estimation, testing, and prediction}, vol.~1.
\newblock Cambridge University Press, 2012.

\bibitem{ulmer2023prior}
D.~T. Ulmer, C.~Hardmeier, and J.~Frellsen, {\it Prior and posterior networks: A survey on evidential deep learning methods for uncertainty estimation},  Transactions on Machine Learning Research (2023)  , \href{http://arxiv.org/abs/2110.03051}{{arXiv:2110.03051 [cs.LG]}}.

\bibitem{van2000asymptotic}
A.~W. Van~der Vaart, {\em Asymptotic statistics}, vol.~3.
\newblock Cambridge university press, 2000.

\bibitem{betancourt2017conceptual}
M.~Betancourt, {\it A conceptual introduction to hamiltonian monte carlo},  arXiv preprint arXiv:1701.02434 (2017)  .

\bibitem{chen2014stochastic}
T.~Chen, E.~Fox, and C.~Guestrin, {\it Stochastic gradient hamiltonian monte carlo},  in {\em International conference on machine learning}, PMLR.
\newblock 2014.

\bibitem{blei2017variational}
D.~M. Blei, A.~Kucukelbir, and J.~D. McAuliffe, {\it Variational inference: A review for statisticians},  Journal of the American statistical Association {\bfseries 112} (2017) 518, 859.

\bibitem{laplace1774memoire}
P.-S. Laplace, {\it Mémoire sur la probabilité des causes par les événemens},  Mémoires de Mathématique et de Physique, Presentés à l'Académie Royale des Sciences, par divers Savans \& lûs dans ses Assemblées {\bfseries 6} (1774)  621.

\bibitem{mackay1992bayesian}
D.~J. MacKay, {\it Bayesian interpolation},  Neural computation {\bfseries 4} (1992) 3, 415.

\bibitem{daxberger2021laplace}
E.~Daxberger, A.~Kristiadi, A.~Immer, R.~Eschenhagen, M.~Bauer, and P.~Hennig, {\it Laplace redux-effortless bayesian deep learning},  Advances in neural information processing systems {\bfseries 34} (2021)  20089.

\bibitem{knoblauch2022an}
J.~Knoblauch, J.~Jewson, and T.~Damoulas, {\it An optimization-centric view on bayes' rule: Reviewing and generalizing variational inference},  Journal of Machine Learning Research {\bfseries 23} (2022) 132, 1, \href{http://arxiv.org/abs/1904.02063}{{arXiv:1904.02063 [stat.ML]}}.

\bibitem{postbayes_seminar}
``Postbayes seminar.'' \url{https://postbayes.github.io/seminar/}.
\newblock Accessed: 2025-11-24.

\bibitem{kernelcalib2019}
D.~Widmann, F.~Lindsten, and D.~Zachariah, {\it Calibration tests in multi-class classification: A unifying framework},  \href{http://arxiv.org/abs/1910.11385}{{arXiv:1910.11385 [stat.ML]}}.

\bibitem{kernelcalib2022}
D.~Widmann, F.~Lindsten, and D.~Zachariah, {\it Calibration tests beyond classification},  \href{http://arxiv.org/abs/2210.13355}{{arXiv:2210.13355 [stat.ML]}}.

\bibitem{Harland-Lang:2024kvt}
L.~A. Harland-Lang, T.~Cridge, and R.~S. Thorne, {\it {A stress test of global PDF fits: closure testing the MSHT PDFs and a first direct comparison to the neural net approach}},  \href{http://dx.doi.org/10.1140/epjc/s10052-025-13934-3}{Eur. Phys. J. C {\bfseries 85} (2025) 3, 316}, \href{http://arxiv.org/abs/2407.07944}{{arXiv:2407.07944 [hep-ph]}}.

\bibitem{DelDebbio:2021whr}
L.~Del~Debbio, T.~Giani, and M.~Wilson, {\it {Bayesian approach to inverse problems: an application to NNPDF closure testing}},  \href{http://dx.doi.org/10.1140/epjc/s10052-022-10297-x}{Eur. Phys. J. C {\bfseries 82} (2022) 4, 330}, \href{http://arxiv.org/abs/2111.05787}{{arXiv:2111.05787 [hep-ph]}}.

\bibitem{1950brier}
G.~W. {Brier}, {\it {Verification of Forecasts Expressed in Terms of Probability}},  \href{http://dx.doi.org/10.1175/1520-0493(1950)078<0001:VOFEIT>2.0.CO;2}{Monthly Weather Review {\bfseries 78} (Jan., 1950)   1}.

\bibitem{sun2018functional}
S.~Sun, G.~Zhang, J.~Shi, and R.~Grosse, {\it Functional variational bayesian neural networks},  in {\em International Conference on Learning Representations}.
\newblock 2019.
\newblock \href{http://arxiv.org/abs/1903.05779}{{arXiv:1903.05779 [cs.LG]}}.

\bibitem{rudner2022tractable}
T.~G. Rudner, Z.~Chen, Y.~W. Teh, and Y.~Gal, {\it Tractable function-space variational inference in bayesian neural networks},  Advances in Neural Information Processing Systems {\bfseries 35} (2022)  22686, \href{http://arxiv.org/abs/2312.17199}{{arXiv:2312.17199 [stat.ML]}}.

\bibitem{gal2016dropout}
Y.~Gal and Z.~Ghahramani, {\it Dropout as a bayesian approximation: Representing model uncertainty in deep learning},  \href{http://arxiv.org/abs/1506.02142}{{arXiv:1506.02142 [stat.ML]}}.

\bibitem{Papamarkou2024}
T.~Papamarkou {\em et al.}, {\it Position: Bayesian deep learning is needed in the age of large-scale ai},  in {\em Forty-first International Conference on Machine Learning}.
\newblock 2024.
\newblock \href{http://arxiv.org/abs/2402.00809}{{arXiv:2402.00809 [cs.LG]}}.

\bibitem{sensoy2018evidentialclass}
M.~Sensoy, L.~Kaplan, and M.~Kandemir, {\it Evidential deep learning to quantify classification uncertainty},  \href{http://arxiv.org/abs/1806.01768}{{arXiv:1806.01768 [cs.LG]}}.

\bibitem{amini2019deepregression}
A.~Amini, W.~Schwarting, A.~Soleimany, and D.~Rus, {\it Deep evidential regression},  \href{http://arxiv.org/abs/1910.02600}{{arXiv:1910.02600 [cs.LG]}}.

\bibitem{meinert2022multivariate}
N.~Meinert and A.~Lavin, {\it Multivariate deep evidential regression},  \href{http://arxiv.org/abs/2104.06135}{{arXiv:2104.06135 [cs.LG]}}.

\bibitem{malinin2018predictive}
A.~Malinin and M.~Gales, {\it Predictive uncertainty estimation via prior networks},  \href{http://arxiv.org/abs/1802.10501}{{arXiv:1802.10501 [stat.ML]}}.

\bibitem{Meinert_2023}
N.~Meinert, J.~Gawlikowski, and A.~Lavin, {\it The unreasonable effectiveness of deep evidential regression},  \href{http://dx.doi.org/10.1609/aaai.v37i8.26096}{Proceedings of the AAAI Conference on Artificial Intelligence {\bfseries 37} (2023) 8, 9134–9142}.

\bibitem{juergens2024}
M.~Jürgens, N.~Meinert, V.~Bengs, E.~Hüllermeier, and W.~Waegeman, {\it Is epistemic uncertainty faithfully represented by evidential deep learning methods?},  \href{http://arxiv.org/abs/2402.09056}{{arXiv:2402.09056 [cs.AI]}}.

\bibitem{DBLP:journals/corr/abs-2106-11642}
F.~D'Angelo and V.~Fortuin, {\it Repulsive deep ensembles are bayesian},  CoRR (2021)  , \href{http://arxiv.org/abs/2106.11642}{{arXiv:2106.11642 [cs.LG]}}.

\bibitem{Kassabov:2022orn}
Z.~Kassabov, M.~Ubiali, and C.~Voisey, {\it {Parton distributions with scale uncertainties: a Monte Carlo sampling approach}},  \href{http://dx.doi.org/10.1007/JHEP03(2023)148}{JHEP {\bfseries 03} (2023)  148}, \href{http://arxiv.org/abs/2207.07616}{{arXiv:2207.07616 [hep-ph]}}.

\bibitem{Costantini:2024wby}
M.~N. Costantini, M.~Madigan, L.~Mantani, and J.~M. Moore, {\it {A critical study of the Monte Carlo replica method}},  \href{http://dx.doi.org/10.1007/JHEP12(2024)064}{JHEP {\bfseries 12} (2024)  064}, \href{http://arxiv.org/abs/2404.10056}{{arXiv:2404.10056 [hep-ph]}}.

\bibitem{rasmussen2006gaussian}
C.~E. Rasmussen and C.~K.~I. Williams, {\em Gaussian Processes for Machine Learning}.
\newblock MIT Press, 2006.

\bibitem{Frate:2017mai}
M.~Frate, K.~Cranmer, S.~Kalia, A.~Vandenberg-Rodes, and D.~Whiteson, {\it {Modeling Smooth Backgrounds and Generic Localized Signals with Gaussian Processes}},  \href{http://arxiv.org/abs/1709.05681}{{arXiv:1709.05681 [physics.data-an]}}.

\bibitem{Horak:2021syv}
J.~Horak, J.~M. Pawlowski, J.~Rodr{\'\i}guez-Quintero, J.~Turnwald, J.~M. Urban, N.~Wink, and S.~Zafeiropoulos, {\it {Reconstructing QCD spectral functions with Gaussian processes}},  \href{http://dx.doi.org/10.1103/PhysRevD.105.036014}{Phys. Rev. D {\bfseries 105} (2022) 3, 036014}, \href{http://arxiv.org/abs/2107.13464}{{arXiv:2107.13464 [hep-ph]}}.

\bibitem{Horak:2023xfb}
J.~Horak, J.~M. Pawlowski, J.~Turnwald, J.~M. Urban, N.~Wink, and S.~Zafeiropoulos, {\it {Nonperturbative strong coupling at timelike momenta}},  \href{http://dx.doi.org/10.1103/PhysRevD.107.076019}{Phys. Rev. D {\bfseries 107} (2023) 7, 076019}, \href{http://arxiv.org/abs/2301.07785}{{arXiv:2301.07785 [hep-ph]}}.

\bibitem{Candido:2024hjt}
A.~Candido, L.~Del~Debbio, T.~Giani, and G.~Petrillo, {\it {Bayesian inference with Gaussian processes for the determination of parton distribution functions}},  \href{http://dx.doi.org/10.1140/epjc/s10052-024-13100-1}{Eur. Phys. J. C {\bfseries 84} (2024) 7, 716}, \href{http://arxiv.org/abs/2404.07573}{{arXiv:2404.07573 [hep-ph]}}.

\bibitem{Medrano:2025cmg}
Y.~C. Medrano, H.~Dutrieux, J.~Karpie, K.~Orginos, and S.~Zafeiropoulos, {\it {Gaussian Processes for Inferring Parton Distributions}},  \href{http://arxiv.org/abs/2510.21041}{{arXiv:2510.21041 [hep-lat]}}.

\bibitem{scikit-learn}
F.~Pedregosa {\em et al.}, {\it Scikit-learn: Machine learning in {P}ython},  Journal of Machine Learning Research {\bfseries 12} (2011)  2825, \href{http://arxiv.org/abs/1201.0490}{{arXiv:1201.0490 [cs.LG]}}.

\bibitem{cover2006elements}
T.~M. Cover and J.~A. Thomas, {\em Elements of Information Theory}.
\newblock Wiley, 2~ed., 2006.

\bibitem{MacKay2003}
D.~J.~C. MacKay, {\em Information Theory, Inference, and Learning Algorithms}.
\newblock Cambridge University Press, 2003.

\bibitem{depeweg2018decomposition}
S.~Depeweg, J.-M. Hernandez-Lobato, F.~Doshi-Velez, and S.~Udluft, {\it Decomposition of uncertainty in bayesian deep learning for efficient and risk-sensitive learning},  in {\em International conference on machine learning}, PMLR.
\newblock 2018.
\newblock \href{http://arxiv.org/abs/1710.07283}{{arXiv:1710.07283 [stat.ML]}}.

\bibitem{houlsby2011bayesian}
N.~Houlsby, F.~Husz{\'a}r, Z.~Ghahramani, and M.~Lengyel, {\it Bayesian active learning for classification and preference learning},  \href{http://arxiv.org/abs/1112.5745}{{arXiv:1112.5745 [stat.ML]}}.

\bibitem{chen2025bayesian}
W.~Chen, B.~Li, R.~Zhang, and Y.~Li, {\it Bayesian computation in deep learning},  \href{http://arxiv.org/abs/2502.18300}{{arXiv:2502.18300 [cs.LG]}}.

\bibitem{VERaiPHY_generative}
S.~Diefenbacher, G.~Kasieczka, and S.~Palacios~Schweitzer, {\it {Generative Models and Statistical Validation}},  to appear (2026)  , \href{http://arxiv.org/abs/2026.xxxxx}{{arXiv:2026.xxxxx}}.

\bibitem{VERaiPHY_robustness}
J.~Cruz-Martinez, C.~Cuesta-Lazaro, A.~Held, and M.~Kagan, {\it {Unknown unknowns in ML for Physics}},  to appear (2026)  , \href{http://arxiv.org/abs/2026.xxxxx}{{arXiv:2026.xxxxx}}.

\bibitem{Ansel2024PyTorch2F}
J.~Ansel {\em et al.}, {\it Pytorch 2: Faster machine learning through dynamic python bytecode transformation and graph compilation},  Proceedings of the 29th ACM International Conference on Architectural Support for Programming Languages and Operating Systems, Volume 2 (2024)  . \href{https://api.semanticscholar.org/CorpusID:268794728}{{\tt https://api.semanticscholar.org/CorpusID:268794728}}.

\bibitem{jax2018github}
J.~Bradbury, R.~Frostig, P.~Hawkins, M.~J. Johnson, C.~Leary, D.~Maclaurin, G.~Necula, A.~Paszke, J.~Vander{P}las, S.~Wanderman-{M}ilne, and Q.~Zhang, {\it {JAX}: composable transformations of {P}ython+{N}um{P}y programs},  version 0.3.13, 2018.
\newblock \href{http://github.com/jax-ml/jax}{{\tt http://github.com/jax-ml/jax}}.

\bibitem{kidger2021equinox}
P.~Kidger and C.~Garcia, {\it {E}quinox: neural networks in {JAX} via callable {P}y{T}rees and filtered transformations},  Differentiable Programming workshop at Neural Information Processing Systems 2021 (2021)  , \href{http://arxiv.org/abs/2111.00254}{{arXiv:2111.00254 [cs.LG]}}.

\bibitem{cabezas2024blackjax}
A.~Cabezas, A.~Corenflos, J.~Lao, and R.~Louf, {\it Blackjax: Composable {B}ayesian inference in {JAX}},  \href{http://arxiv.org/abs/2402.10797}{{arXiv:2402.10797 [cs.MS]}}.

\end{thebibliography}\endgroup
\end{document}